\documentclass[10pt,twocolumn,letterpaper]{article}

\usepackage{cvpr}      
\usepackage[accsupp]{axessibility}

\usepackage{multirow} 

\definecolor{cvprblue}{rgb}{0.21,0.49,0.74}
\usepackage[pagebackref,breaklinks,colorlinks,allcolors=cvprblue]{hyperref}

\def\confYear{2026}

\title{StyleGallery: Training-free and Semantic-aware Personalized Style Transfer \\from Arbitrary Image References}

\author{\textbf{Boyu He$^{1*}$, Yunfan Ye$^{2*}$, Chang Liu$^{1}$, Weishang Wu$^{1}$, Fang Liu$^{2\dagger}$, Zhiping Cai$^{1\dagger}$} \\
\small $^1$College of Computer Science and Technology, National University of Defense Technology\\
\small $^2$School of Design, Hunan University
}

\usepackage{tabularx}
\usepackage{algorithm}
\usepackage{algorithmic}
\usepackage{amsmath}
\begin{document}

\maketitle
\def\thefootnote{*}\footnotetext{Equal contribution}
\def\thefootnote{$\dagger$}\footnotetext{Corresponding authors}
\begin{abstract}
Despite the advancements in diffusion-based image style transfer, existing methods are commonly limited by 1) semantic gap: the style reference could miss proper content semantics, causing uncontrollable stylization; 2) reliance on extra constraints (e.g., semantic masks) restricting applicability; 3) rigid feature associations lacking adaptive global-local alignment, failing to balance fine-grained stylization and global content preservation. 
These limitations, particularly the inability to flexibly leverage style inputs, fundamentally restrict style transfer in terms of personalization, accuracy, and adaptability.
To address these, we propose StyleGallery, a training-free and semantic-aware framework that supports arbitrary reference images as input and enables effective personalized customization.
It comprises three core stages: semantic region segmentation (adaptive clustering on latent diffusion features to divide regions without extra inputs); clustered region matching (block filtering on extracted features for precise alignment); and style transfer optimization (energy function-guided diffusion sampling with regional style loss to optimize stylization).
Experiments on our introduced benchmark demonstrate that StyleGallery outperforms state-of-the-art methods in content structure preservation, regional stylization, interpretability, and personalized customization, particularly when leveraging multiple style references. The source code and dataset are available at \url{https://github.com/iiiiiiiword/StyleGallery}.
\end{abstract}    
\section{Introduction}
\label{sec:intro}

Style transfer technology~\cite{huang2017arbitrary, Gu2018ArbitraryST, liu2021adaattn, deng2022stytr2, Chung2023StyleII}, with its ability to blend arbitrary stylistic elements into content images and yield striking visual transformations, has garnered notable attention in recent years. This owes much to its diverse applications spanning graphic design and digital art, user-specific personalization, and beyond. 

Recent advances in Diffusion Models (DMs)~\cite{Ramesh2022HierarchicalTI, zhang2023inversion} have also led to breakthroughs in style transfer, such as training-free frameworks~\cite{Chung2023StyleII, zhou2025attention} that achieve stable transfer by modifying components within the self-attention module of pre-trained DMs, or other effort~\cite{shang2025scsa} that achieves strong semantic matching in style transfer by inputting extra semantic masks.

Despite the favorable achievements, current paradigms still struggle to achieve a balance between style transfer and content preservation. For example, while StyleID~\cite{Chung2023StyleII} demonstrates excellent content preservation through direct manipulation of attention features, it suffers from insufficient stylization, even producing random patterns when faced with solid-color backgrounds (Figure~\ref{fig:Zoomed-compare}).
Attention Distillation~\cite{zhou2025attention} (AD) introduces an energy function to constrain the denoising direction, achieving progress in stylization but introducing the risk of content leakage~\cite{an2021artflow, ye2025stylemaster, lin2025inversion}. 
From above, we conclude three typical types of challenges:
1) Semantic gap: the semantic information of a single style reference may fail to cover the semantics of the content image (e.g., the content image contains ``mountains" but the style image does not), resulting in unreasonable stylization regions in handling inputs with a semantic gap. 
2) Extra constraint: Additional semantic masks are required~\cite{shang2025scsa}, and the semantic structures between content and style images are constrained to be similar, limiting the applications.
3) Rigid features: Existing methods lack adaptive global-regional semantic association between content and style features~\cite{Xing2024CSGOCC, gao2025styleshot}, failing to identify optimal styles for fine-grained transfer. Rigid features without adaptive matching limit handling of semantic discrepancies, constraining regional stylization and ultimately falling short of meeting the growing demand for personalized customization.

In this work, we still follow the assumption that semantic regions are the fundamental carriers of style characteristics, and adaptive matching between content and style regions is critical for high-quality style transfer~\cite{shang2025scsa}. To tackle the aforementioned challenges, we propose StyleGallery, a training-free and semantic-aware style transfer framework that enables aggregating an arbitrary number of style references through adaptive feature clustering and matching, and optimizing via global and regional loss. The main difference between StyleGallery and the general style transfer is shown in Figure~\ref{fig:teaser}. Such a paradigm also benefits practical applications like serial style transfer (e.g., from all works of a specific artist) and personalized transfer (with user-defined choices).

Specifically, we first perform semantic region segmentation by extracting weighted intermediate features from the UNet~\cite{ronneberger2015u} and use K-means~\cite{Ikotun2022KmeansCA} clustering to segment regions, thereby avoiding external models for flexibility.
For semantic region matching, we employ three dimensions, with particular emphasis on the semantic match: DINOv2~\cite{Oquab2023DINOv2LR} is used to extract region-level features and measure similarity, enabling adaptive alignment of content regions with the most relevant style regions from multiple references.
For semantic-aware style transfer, inspired by the energy constraint from AD~\cite{zhou2025attention}, we propose a regional style loss to guide diffusion sampling via classifier guidance~\cite{dhariwal2021diffusion}, enhancing regional style consistency while preserving global structural coherence.

Extensive experiments on the collected dataset demonstrate that StyleGallery outperforms existing methods both with single or multiple input references. The user studies also validate our advantages in content preservation and stylization. Our contributions mainly include:

\begin{itemize}
\item A training-free and semantic-aware framework for arbitrary image reference style transfer, offering a new paradigm for achieving personalized style transfer.

\item Several technical designs of adaptive clustering, matching and optimization, to prevent insufficient stylization and content leakage.

\item A dataset containing various style galleries (a series of style images) for evaluating and comparing style transfer with multiple input references.
\end{itemize}
\section{Related Work}
\label{sec:related-work}
\subsection{Neural Style Transfer via Diffusion Models}
Neural style transfer~\cite{Gatys2016ImageST,Yi2017neural,Lu2019ACS,liu2021adaattn,deng2022stytr2,Zhang2022DomainEA,wu2022ccpl} aims to transfer the style of one image to another while preserving the original content information, and has successfully facilitated style transfer through the fusion of style and content information.
With the rise of DMs, this field has gradually introduced their generative capabilities to achieve better style transfer results. DiffStyle~\cite{Jeong2023TrainingfreeST} provides a novel, training-free approach via the h-space~\cite{Kwon2022DiffusionMA}, and CSGO~\cite{Xing2024CSGOCC} offers a seamless, end-to-end style transfer that blends style and content without additional fine-tuning. StyleShot~\cite{gao2025styleshot} uses a sophisticated encoder for nuanced style embeddings, supporting complex transfers from diverse inputs. Although these methods perform well, they still struggle to achieve interpretable style transfer while maintaining the original spatial structure and semantic information.

In this paper, we aim to achieve semantic-aware region-level style transfer while ensuring content preservation, enhancing the interpretability of the transfer process. We do not treat style as a single holistic feature to be transferred, but adopt a divide-and-conquer strategy to transfer regional style features based on semantic relevance, which not only improves the interpretability and rationality of the results but also supports personalized customization.

\subsection{DM-Based Attention Editing}
Numerous works manipulate DM-based attention for editing the generation process and results. 
Pre-trained text-to-image (T2I) DM~\cite{Ramesh2022HierarchicalTI} has significantly advanced image editing techniques~\cite{Avrahami2021BlendedDF, Shi2023DragDiffusionHD}. 
Prompt-to-Prompt~\cite{Hertz2022PrompttoPromptIE} enables text-based image editing through cross-attention maps, linking spatial distribution in the image to text cues. 
StyleID~\cite{Chung2023StyleII} utilizes DDIM Inversion to extract content queries ($Q$), style keys ($K$) and values ($V$) from the image, and injects them into the self-attention layers of UNet~\cite{ronneberger2015u} for style transfer.
Attention Distillation (AD)~\cite{zhou2025attention} introduces a loss function to transfer visual features from a reference image to a generated one by optimizing latent features. 
However, these works primarily focus on global stylistic features of images, neglecting the importance of semantic correspondence in style transfer, leading to inconsistencies and unintended styles in generated images.

To alleviate this concern, we perform adaptive region clustering and matching for the content and style images at the level of DM-based attention features, which can thus achieve reasonable and coherent stylization results.

\subsection{Cross-Image Semantic Matching}
Semantic correspondence involves matching regions with identical semantics across different images. Traditional methods~\cite{Zhang2021DatasetGANEL, Zhao2021MultiscaleMN} rely on handcrafted features with complex processing, while CNN-based deep learning approaches, such as Mask R-CNN~\cite{he2017mask}, offer end-to-end learning but at higher computational costs. DMs paired with modern vision models such as DINOv2~\cite{Oquab2023DINOv2LR} and CLIP~\cite{Radford2021LearningTV} demonstrate strong generalization but still face challenges in semantic region classification and matching.
Semantix~\cite{He2025SemantixAE} achieves point-level semantic matching based on DIFT~\cite{tang2023emergent}, but it focuses only on limited point correspondences. 
SCSA~\cite{shang2025scsa} can attain region-level semantic matching by adding semantic segmentation maps, but relies on extra mask inputs and strong semantic correspondences between styles and content images. 

In this work, we compute similarities across three dimensions to achieve adaptively optimal matching. Without requiring any additional input, our method enables accurate semantic region matching across diverse images.

\section{Method}
\label{sec:method}
\begin{figure*}[t] 
    \centering
    \includegraphics[width=0.85\linewidth]{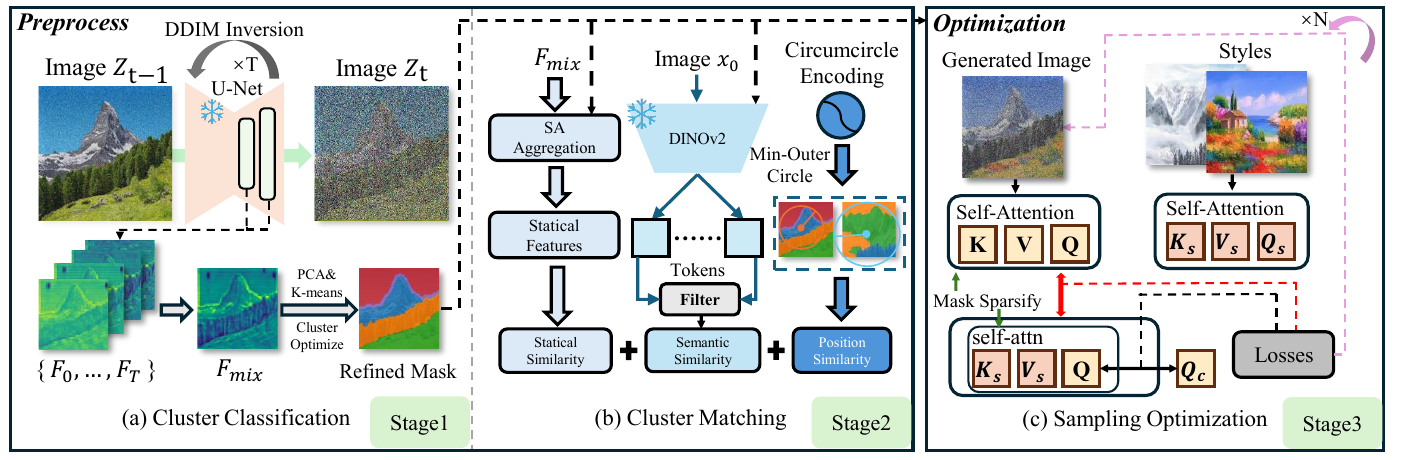} 
    \caption{
    Overall framework. Our pipeline comprises three stages: (a) In stage 1, the content image is diffused for T steps to extract UNet features ${F_0,\dots,F_T}$, which are weighted into $F_{mix}$, then clustered via PCA and K-means to optimize the mask. (b) In stage 2, $F_{mix}$ and the semantic mask identify cluster features, which are aggregated via self-attention for statistical similarity. Meanwhile, DINOv2~\cite{Oquab2023DINOv2LR} splits $x_0$ into blocks; tokens are filtered for semantic similarity, while cluster positions yield positional similarity. (c) Stage 3 optimizes the generation through $N$ latent sampling steps. UNet attention maps are extracted, sparsified using the semantic mask, and recombined with style features ($K_s$, $V_s$). L1 distances are then computed between the actual and combined feature maps, as well as the $Q$ and $Q_c$. These losses guide the final image generation. }
    \label{fig:pipeline}
\end{figure*}

\begin{figure}[ht]
    \centering
    \includegraphics[width=0.9\linewidth]{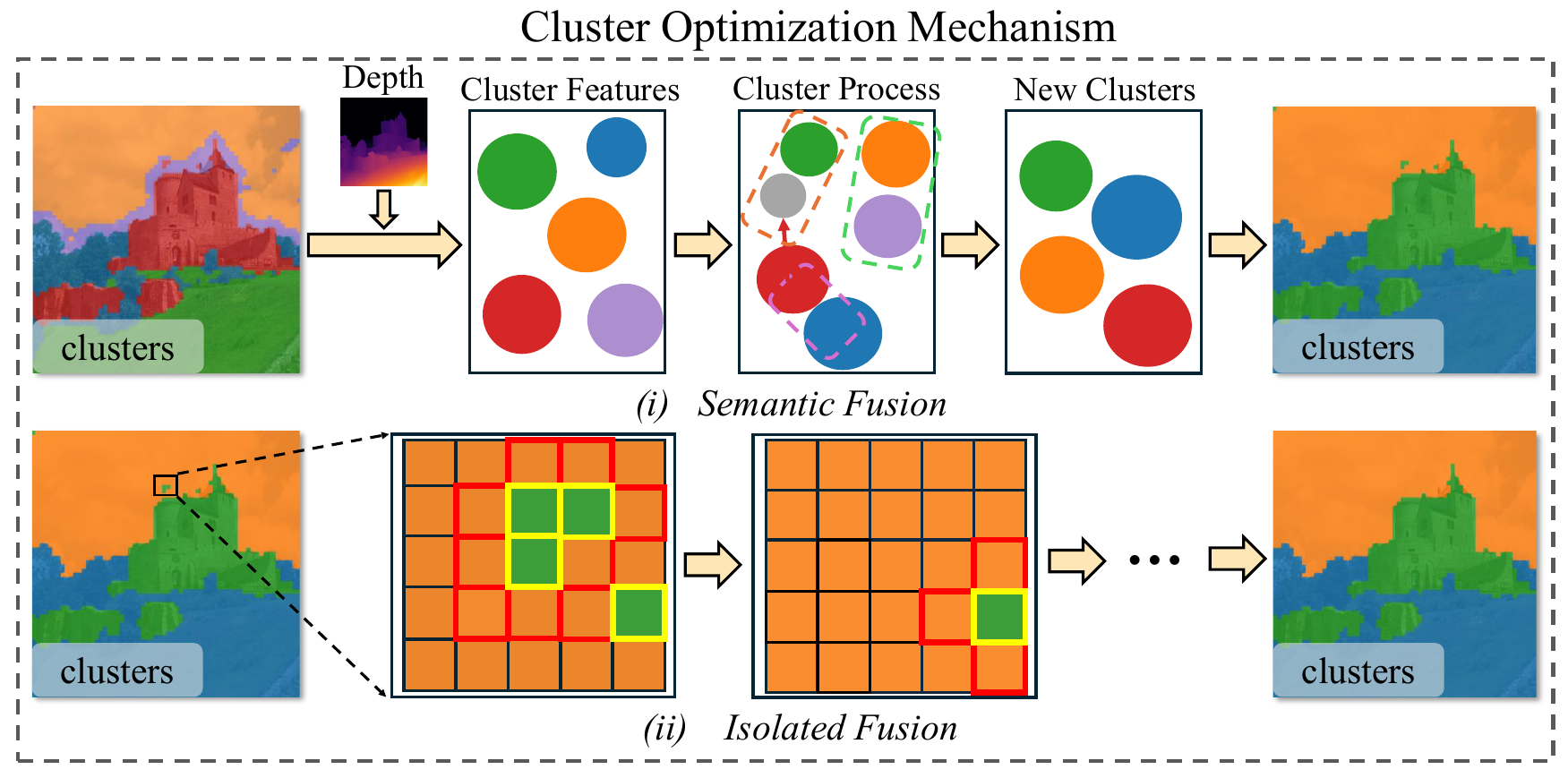}
    \caption{Cluster optimization. 
    We compute pairwise semantic distances among clusters, merge those below a threshold (set to 0.85), then split–recombine clusters guided by the input’s depth features, then traverse each pixel, eliminate isolated points.}
    \label{fig:cluster-optimize}
\end{figure}

\subsection{Preliminaries}
\paragraph{Latent Diffusion Models.}
LDM~\cite{rombach2022high} is a state-of-the-art generative model that operates in a latent space to significantly reduce computational overhead. Given an input image, a pretrained VAE encodes it into a learned latent representation. Then a denoising network ${\epsilon }_{\theta}(\cdot)$ based on UNet~\cite{ronneberger2015u} is trained to predict the noise at each step of the diffusion process by minimizing the mean squared error between the predicted noise ${\epsilon}_{\theta}$ and the actual added noise $\epsilon$. The objective function is as follows:
\begin{equation}
\mathcal{L}_{\text{LDM}} = \mathbb{E}_{z , y, \epsilon, t} \left[ \left\| \epsilon -\epsilon_{\theta}(z_t, t, y)\right\|_2^2 \right] 
\end{equation}
where $\epsilon\in\mathcal{N}(0,1)$ represents the noise, $t$ denotes the timestep and $y$ is a condition.
\paragraph{Attention Mechanisms in Diffusion.}
Self-Attention (SA) and Cross-Attention (CA) mechanisms are pivotal in the UNet-based~\cite{ronneberger2015u} diffusion architecture. Many studies~\cite{Hertz2022PrompttoPromptIE, Tumanyan2022PlugandPlayDF, Deng2023ZZS, gu2024swapanything} have demonstrated that attention maps exhibit strong information integration capabilities, enabling them to capture information related to image structure and content.
We focus on the SA module for style transfer, and the self-attention mechanism can be formulated as:
\begin{equation}
\text{Self-Attn}(Q, K, V) = \text{Softmax}\left( \frac{QK^T}{\sqrt{d}} \right)V 
\end{equation}
where $Q$, $K$, and $V$ represent visual features and $d$ is the feature dimension used for scaling.

\paragraph{Diffusion Features.}
DIFT~\cite{tang2023emergent} is an unsupervised correspondence extraction framework that leverages a pretrained diffusion model. It adds controlled Gaussian noise into input images and extracts intermediate UNet activations to generate pixel-level representations, achieving cross-view, cross-category, and cross-modal consistency.

Building on this, we apply semantic-aware K-means~\cite{Ikotun2022KmeansCA} clustering to intermediate UNet feature maps to identify distinct semantic regions. We then perform semantic decomposition on the attention features, processing the features associated with each region separately to achieve semantically aligned regional style transfer.

\subsection{Pipeline Overview} 
Given a style set $I_s = {I_1, I_2, \ldots}$ and a content image $I_c$, StyleGallery adaptively identifies semantically matching regions from $I_s$ for stylization while preserving content structure and reducing leakage.
As shown in Figure~\ref{fig:pipeline}, we first use Diffusion Features for Cluster Classification (DFCC) to perform DDIM inversion over $T$ timesteps on both images, extracting and fusing UNet intermediate features into a unified feature. K-means clustering then identifies semantic regions, with masks refined via cluster optimization.
Then, region-wise similarity is computed along three dimensions: statistical similarity from U-Net feature statistics, semantic similarity from region features extracted by DINOv2~\cite{Oquab2023DINOv2LR}, and positional similarity from the minimum enclosing circles. 
During $N$ sampling steps, self-attention features from UNet~\cite{ronneberger2015u} are used to minimize the L1 distance between real and ideal feature maps, guiding the final generation.

\subsection{Diffusion Features for Cluster Classification}
Recent style transfer works~\cite {shang2025scsa, He2025SemantixAE} focus on semantic correspondences between content and style images for controllable generation. For example, Semantix~\cite{He2025SemantixAE} achieves this through pixel-to-pixel matching, while SCSA~\cite{shang2025scsa} relies on semantic segmentation maps.

Unlike these works, we adopt a training-free semantic clustering strategy based on DIFT~\cite{tang2023emergent}. We input an image into a pretrained diffusion model, extract UNet intermediate feature maps via forward diffusion, then utilize these maps to differentiate semantic regions. The features are then used for cluster matching in the next stage. 
The method can be divided into four steps:
1) Extract feature maps from UNet~\cite{ronneberger2015u} for semantic clustering and obtain the initial noisy feature map;
2) Apply index-adaptive weighting to the feature map sequence to get $F_{mix}$;
3) Perform PCA dimensionality reduction and K-means clustering on $F_{mix}$;
4) Cluster Optimization (shown in Figure~\ref{fig:cluster-optimize}) based on (i) semantic similarity (ii) fusing isolated point clusters with their nearest neighbors. 
Step two can be formulated as:
\begin{equation}
d(t) = \frac{1}{1 + \exp\left( 5 \cdot \left(\frac{t}{T} - 0.7 \right) \right)}
\end{equation}
\begin{equation}
F_{mix} = \sum_{t}^{T} 
\left(\frac{d(t)}{\sum_{k}^{T} d(k)}\right)\cdot F_t 
\end{equation}
where $t$ is the current timestep, $T$ the total number of timesteps, and $F_t$ the UNet output at timestep $t$. The hyperparameters 5 and 0.7 control the steepness and the inflection point of the weight curve, respectively. The whole process of DFCC can be formulated as:
\begin{equation}
\text{Clusters} = Optimization(\text{K-means}\left( F_{mix}, K \right), F) 
\end{equation}
where $F$ represents the initial image feature after VAE encoding, and $K$ denotes the upper limit of the number of cluster classifications.

\subsection{Cluster Matching}
Since content and style images differ in style, color, shape, and texture, we model these variations and align corresponding semantic regions by computing similarities across three feature dimensions for adaptively optimal matching:

Dimension 1 (Statistical Features): Apply a cluster mask to the fused feature map to isolate features from different semantic regions. For points within the same region, we use self-attention to aggregate relationships and compute statistical features (mean, variance) for each cluster.

Dimension 2 (Semantic Similarity): Input the image into DINOv2~\cite{Oquab2023DINOv2LR} to obtain semantic feature tokens and reshape the cluster mask to match these tokens. Concatenate tokens if the valid count exceeds 100 to form cluster features, and compute cosine similarity for semantic matching.

Dimension 3 (Geometric Criterion): When semantic correspondence between the content and style images is weak, we introduce cluster outer circle similarity. We compute the minimum enclosing circle for each cluster to capture position info (center and radius), providing additional data for matching.
Therefore, the final formula we defined for semantic matching is expressed as:
\begin{equation}
\text{Similarity} = \sum_{i}^{} {\lambda_i} * CS({feat_i^c}, feat_i^s) 
\end{equation}
where $\lambda_i$ is the weight of the $i$-th dimension, and $feat_i^a$, $feat_i^b$ are features from the content and style image clusters obtained via semantic clustering masks. 
$CS(a,b)$ denotes the cosine similarity between vector $a$ and $b$. 
We set $\lambda_1=0.25$, $\lambda_2=1$, and $\lambda_3=0.125$.

\subsection{Sampling Optimization}
\begin{figure}[t] 
    \centering
    \includegraphics[width=0.85\linewidth]{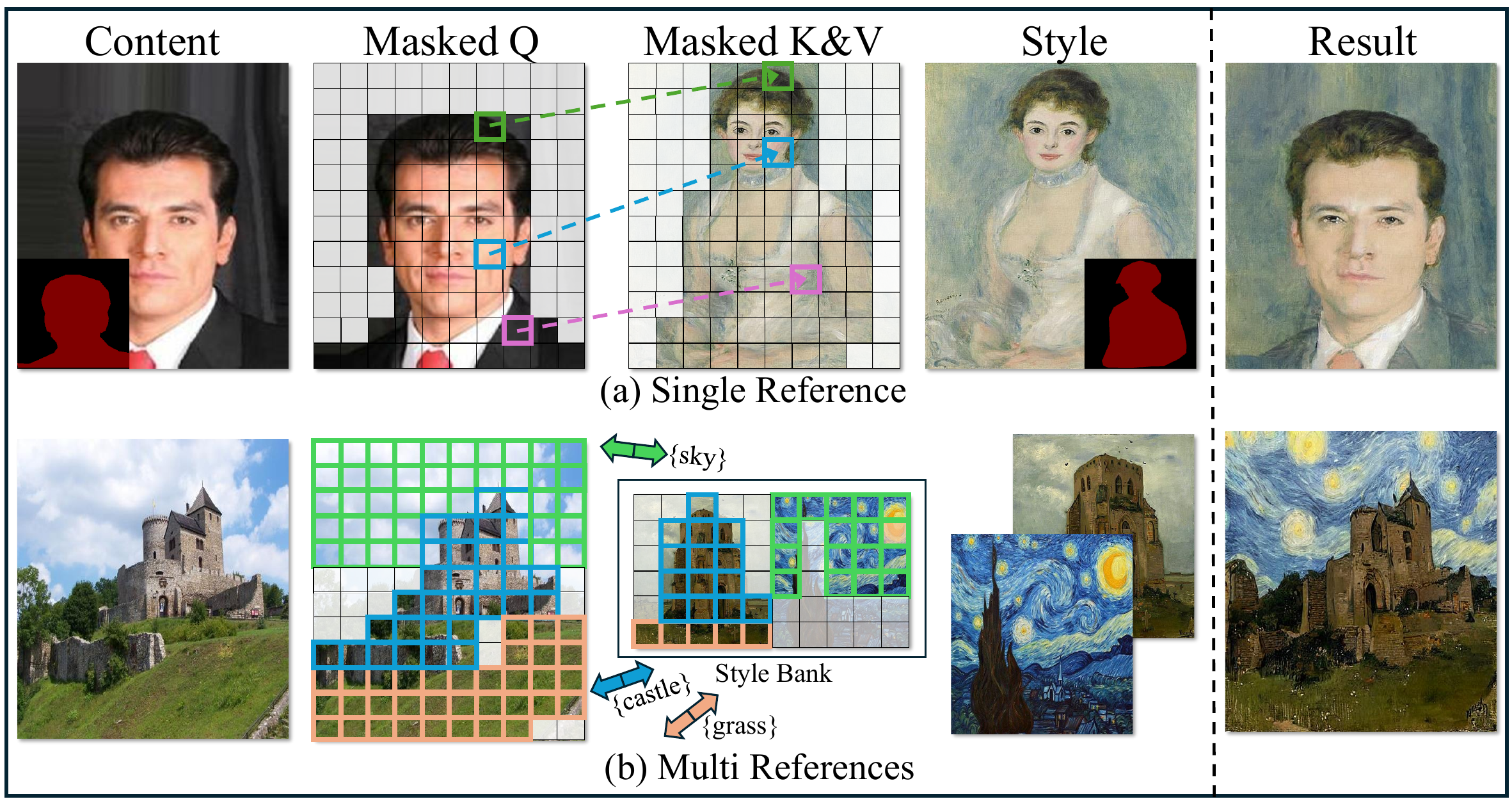}
    \caption{Sparse attention based on semantic matches. 
    We mask the attention feature map, preserving weights for relevant semantics and zeroing out irrelevant regions. Content-sparse attention matches corresponding style features, yielding optimal stylization. 
    Semantically identical regions are color-coded; connections use matching colors.
    }
    \label{fig:attn}
\end{figure}

After establishing semantic region matching, transferring regional style features to the corresponding regions in the content image is the third problem addressed in our work. Inspired by prior works such as AD~\cite{zhou2025attention} and Semantix~\cite{He2025SemantixAE}, we adopt two losses: regional style loss and global content loss.
The details of these losses are as follows:

(1) Regional Style Loss (RSL):
We extract $Q$, $K$, $V$ features from the last 6 self-attention layers of UNet. With the semantic cluster mask, we sparsify features for each cluster by zeroing out unrelated points and reshaping the mask to align with the dimensions of $Q$, $K$, $V$. This feature weight mask is then applied to $Q$, $K$, $V$ to preserve weights for the related semantic region and nullify others as shown in Figure ~\ref{fig:attn}. Finally, we compute the L1 loss between each pair of matched semantic regions and aggregate it as:
\begin{equation}
\begin{aligned}
\mathcal{L}_{\text{RSL}}
&= \sum_{i,j}\Bigl\lVert
     \text{Mask}\bigl(\text{Self-Attn}(Q_i,K_i,V_i)\bigr) \\
&\quad - \text{Self-Attn}\bigl(\text{Mask}(Q_i),
     \text(K^s_j),\text(V^s_j)\bigr)
   \Bigr\rVert_1
\end{aligned}
\end{equation}
where i, j denotes the cluster index from the generated and style images respectively.

(2) Global Content Loss (GCL): To maintain content consistency between the generated and the original image, we use the content loss from AD~\cite{zhou2025attention} as our global content loss, which can be expressed as, where $Q$ is obtained from the generated branch and $Q_c$ from the content branch:
\begin{equation}
\mathcal{L}_{\text{GCL}} = \left\lVert Q - Q_c \right\rVert_1
\end{equation}
Thus we can get the Overall loss:
\begin{equation}
\mathcal{L}_{\text{RST}} = \mathcal{L}_{\text{RSL}} + \lambda_c *\mathcal{L}_{\text{GCL}}
\end{equation}
where $\lambda_c$ is the percentage of content loss that can be adjusted to control the generation result.

We interpret $\mathcal{L_\text{RST}}$ as an energy function to guide the DDIM sampling process, then a gradient-based optimizer (Adam~\cite{kingma2014adam} in practice) updates the latent vector $z_{t-1}$ according to:
\begin{equation}
z_{t-1} = z_{t-1} - \eta\nabla_{z_{t-1}} L_{RST}(z_{t-1}, z_{t-1}^{\text{ref}})
\end{equation}
where $z_{t-1}$ denotes the latent vector at timestep $t-1$, $z_{t-1}^{ref}$ is the style latent perturbed by noise, $\eta$ is set to 0.05.

\begin{figure*}[t] 
    \centering
    \includegraphics[width=0.9\linewidth]{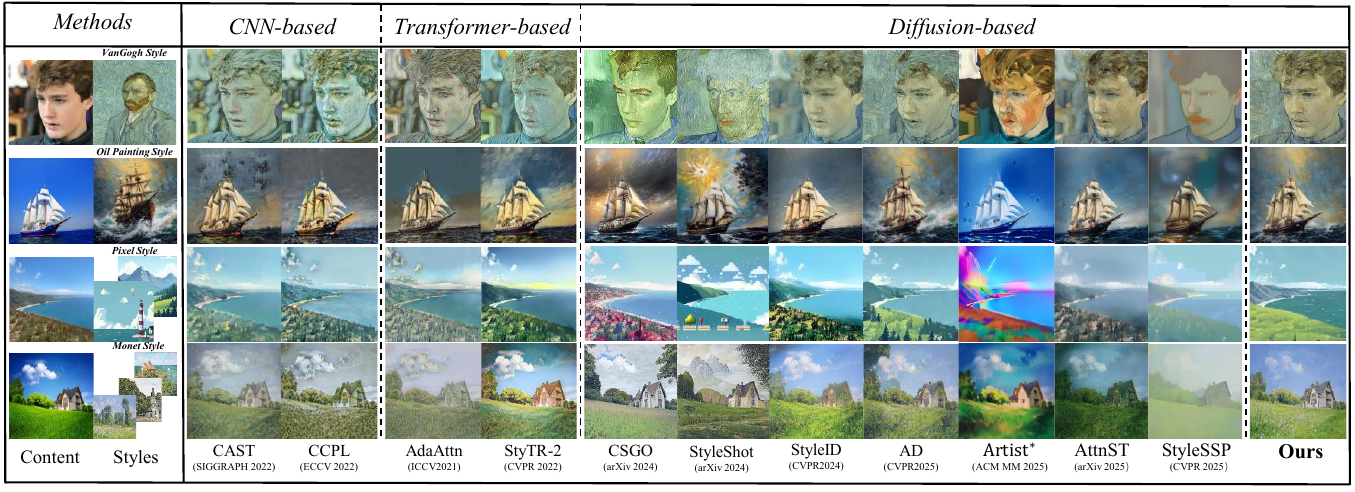} 
    \caption{Qualitative comparisons with recent state-of-the-art image style transfer methods. 
    Our StyleGallery achieves semantic-level style transfer for fine-grained content (e.g., hair and clothing in the first row; sea and sailboat in the second row) while preserving content consistency, and others either fail to transfer styles accurately or compromise content structure. Regarding multiple references, our method can accurately and adaptively recognize, match, and transfer regional styles (the third row) without introducing incorrect semantic content seen in others (e.g., unexpected semantics in the sky of other methods in the fourth row).  
    Zoom-in is recommended for better comparisons.}
    \label{fig:qualitative-compare}
\end{figure*}

\section{Experiments}
\label{sec:Experiments}
\subsection{Experimental Settings}
We conducted all experiments using pretrained Stable Diffusion 1.5~\cite{rombach2022high} on an NVIDIA RTX 4090. The forward diffusion process consisted of 15 steps
, with optimization performed over 150 steps during image generation. 
Unless noted, default hyperparameters are: upper-limit on semantic clustering with (K=10); semantic-similarity weight ($\lambda_2=1$) with the remaining loss terms kept in the ratio (2{:}8{:}1); global content-loss weight ($\lambda_c=0.26$).

\textbf{Dataset.}
Since there are no benchmarks specifically designed for style transfer from arbitrary image reference, for evaluation, we collect content and style images sourced from the Internet and public datasets, including COCO~\cite{Lin2014MicrosoftCC}, FFHQ~\cite{Karras2017ProgressiveGO} and WikiArt~\cite{Tan2017ImprovedAF}. The set covers 25 style families spanning different periods, regions, and artists (e.g., Van Gogh and Chinese ink), with 4-17 images per style (mean=8). Content images are grouped into five categories, each containing 2-10 images. For each content image, we randomly select 1, 2, 3, or 5 style images per category, resulting in a total of 750 stylized images.

\begin{table*}[t]
    \centering
    \renewcommand{\arraystretch}{1.0} 
    \setlength{\tabcolsep}{3pt}
    \scriptsize
    \resizebox{0.8\textwidth}{!}{%
    \begin{tabular}{l|cc|cc|ccccc|c}
        \toprule  
        Metric & CAST~\cite{Zhang2022DomainEA} & CCPL~\cite{wu2022ccpl} & AdaAttn~\cite{liu2021adaattn} & StyTR-2~\cite{deng2022stytr2} & CSGO~\cite{Xing2024CSGOCC} & StyleShot~\cite{gao2025styleshot} & StyleID~\cite{Chung2023StyleII} & AD~\cite{zhou2025attention} & AttnST~\cite{Huang2025AttenSTAT} & \textbf{Ours} \\
        \midrule  
        Style ↑ & 0.4959 & 0.5163 & 0.5094 & 0.5219 & 0.5224 & 0.5198 & 0.4972 & \underline{0.5249} & 0.5032 & \textbf{0.5337} \\
        Gram Loss ↓ & 15.778 & 17.8606 & 17.357 & 16.719 & 19.937 & 19.013 & 14.261 & \underline{13.862} & 16.937 & \textbf{13.519}\\
        \midrule  
        FID~\cite{heusel2017gans} ↓ & \underline{17.255} & 18.141 & 18.498 & 17.623 & 19.829 & 20.638 & 18.987 & 17.677 & 19.233 & \textbf{16.889} \\
        LPIPS~\cite{Zhang2018TheUE} ↓ & 0.4934 & 0.4549 & 0.5299 & \underline{0.3856} & 0.5005 & 0.6615 & 0.4496 & 0.4032 & 0.4532 & \textbf{0.3716} \\
        ArtFID~\cite{wright2022artfid} ↓ & 27.262 & 27.843 & 29.831 & \underline{25.804} & 31.254 & 35.952 & 28.973 &
        26.207 & 29.402 & \textbf{24.536}\\
        \bottomrule  
    \end{tabular}}
    \caption{Quantitative comparison with image style transfer methods. The best results are shown in \textbf{bold}, and the second-best results are indicated with an \underline{underline}.}  
    \label{tab:quantity-comparison}
\end{table*}

\begin{figure}[t]
    \centering
    \includegraphics[width=0.85\linewidth]{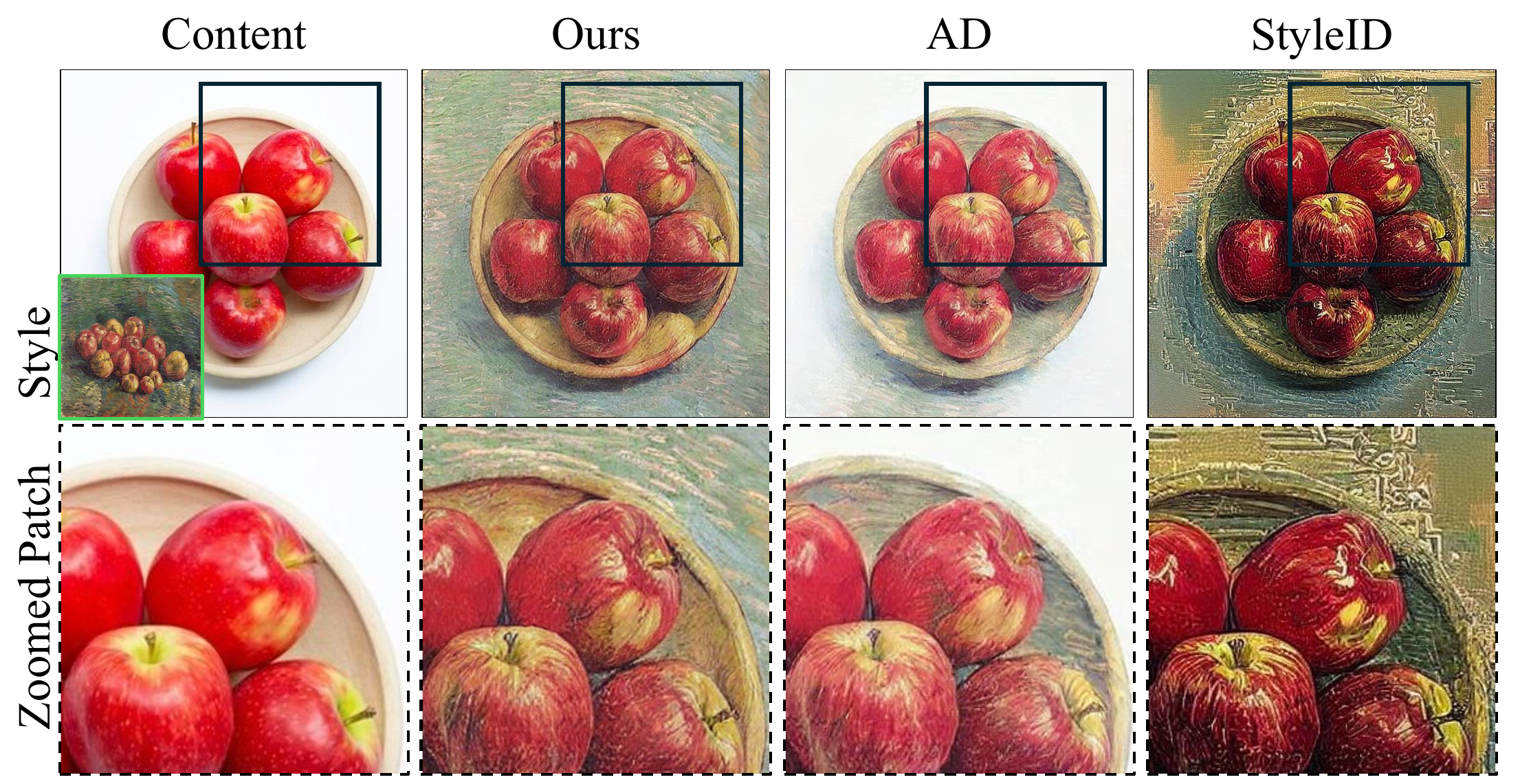}
    \caption{Zoomed details of our method, AD~\cite{zhou2025attention} and StyleID~\cite{Chung2023StyleII}. Black boxes highlight enlarged views of the results.AD~\cite{zhou2025attention} under-stylizes the image, with the background remaining unchanged, while StyleID~\cite{Chung2023StyleII} introduces incorrect patterns. This underscores the need for our semantic-aware regional style transfer.}
    \label{fig:Zoomed-compare}
\end{figure}

\textbf{Metrics.}
As our method accommodates any number of style images, for a fair comparison to evaluate style transfer quality, we adopt a metric based on the Hungarian algorithm~\cite{Kuhn1955TheHM} by splitting images into 16 non-overlapping 128×128 blocks. A VGG network extracts texture, color, and style features, and each content block is matched to its nearest style block using cosine distance. The average style distance across these matches is computed as our style metric, termed ``Style'' in Table~\ref{tab:quantity-comparison}.
We additionally compute Gram matrices from these block features, and perform block matching based on L1 distance to report the average ``Gram'' loss, which are widely used in style transfer~\cite{Gatys2016ImageST, Alaluf2023CrossImageAF}. 
Furthermore, Fréchet Inception Distance (FID)~\cite{heusel2017gans} is used to assess style fidelity between the stylized and style images, while LPIPS~\cite{Zhang2018TheUE} quantifies structural consistency between the content and stylized images. We also compute ArtFID~\cite{wright2022artfid} (ArtFID=(1+FID)·(1+LPIPS)), which evaluates overall style transfer quality with considering both content and style preservation.

\begin{figure}[t]
    \centering
    \includegraphics[width=1.0\linewidth]{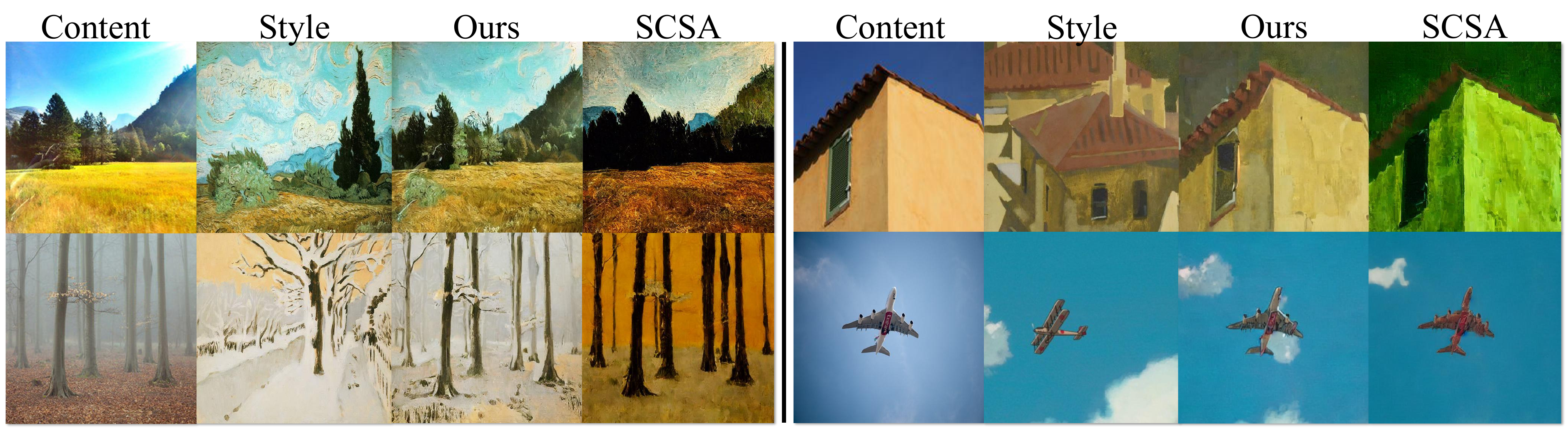}
    \caption{Comparisons with mask-based method SCSA~\cite{shang2025scsa}.}
    \label{fig:scsa}
\end{figure}

\begin{figure*}[ht]
    \centering
    \includegraphics[width=0.87\linewidth]{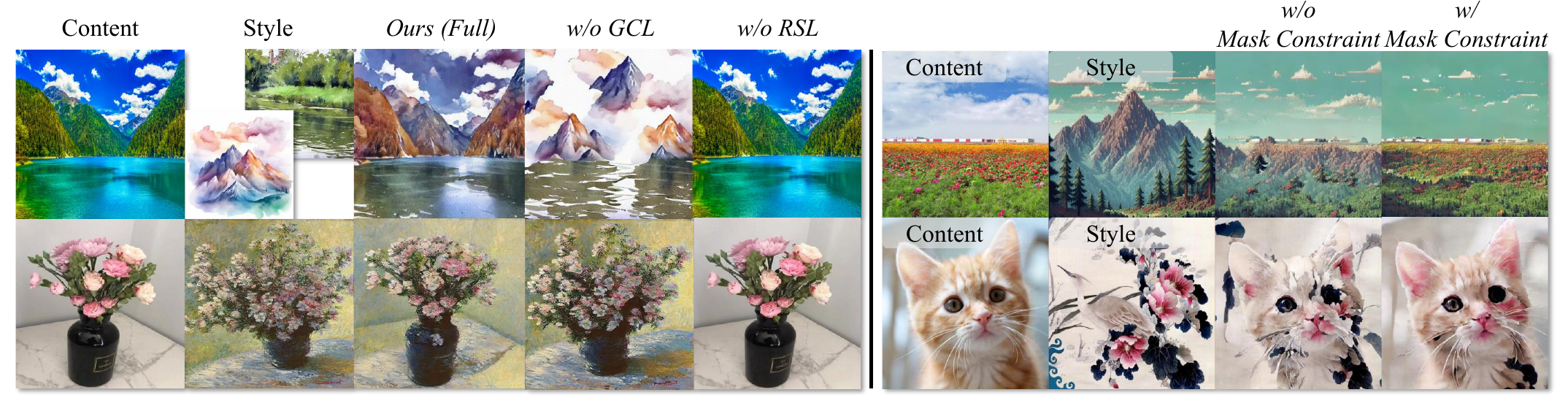}
    \caption{Ablation study on loss functions. The global content loss preserves semantic consistency while the region style loss enables effective transfer, with their combination yielding faithful stylization; additionally, confining style transfer to semantically matched region pairs further enhances quality and mitigates semantic leakage in the generated images.}
    \label{fig:ab-1}
\end{figure*}

\subsection{Comparison Results} 
We implement a personalized semantic corresponding style transfer method for arbitrary image inputs and compare it with previous state-of-the-art methods: CNN-based methods CAST~\cite{Zhang2022DomainEA}, CCPL~\cite{wu2022ccpl}; Transformer-based methods AdaAttn~\cite{liu2021adaattn}, StyTr-2~\cite{deng2022stytr2}; Diffusion-based methods CSGO~\cite{Xing2024CSGOCC}, StyleShot~\cite{gao2025styleshot}, StyleID~\cite{Chung2023StyleII}, AttentionDistillation~\cite{zhou2025attention} (AD), Artist~\cite{jiang2025diffartist}, AttnST~\cite{Huang2025AttenSTAT} and StyleSSP~\cite{Xu2025StyleSSPSS}. 
More results and discussion, including the user study, can be found in the Supplementary.

\textbf{Qualitative Comparisons.}
Unlike single-reference methods such as StyleID~\cite{Chung2023StyleII}, StyleShot~\cite{gao2025styleshot}, and CSGO~\cite{Xing2024CSGOCC}, our approach natively supports multiple style images.
For CNN-based and Transformer-based methods such as CCPL~\cite{wu2022ccpl} and AdaAttn~\cite{liu2021adaattn}, we average features across the selected references before feeding them to the model. For diffusion-based methods, StyleShot~\cite{gao2025styleshot}, CSGO~\cite{Xing2024CSGOCC}, AttnST~\cite{Huang2025AttenSTAT} and StyleSSP~\cite{Xu2025StyleSSPSS} follow a similar approach by extracting and concatenating style features. For StyleID~\cite{Chung2023StyleII}, we sample attention features from each style image and concatenate them.
AttentionDistillation (AD)~\cite{zhou2025attention} inherently supports multi-style inputs, requiring no modifications. For Artist~\cite{jiang2025diffartist}, we use ChatGPT to generate a comprehensive description of the style.

Figure~\ref{fig:qualitative-compare} presents the qualitative comparison results. Our method effectively transfers styles from matching semantic regions of style images to the content image, achieving high-quality semantically-aware style transfer, particularly evident in the Oil Painting and Pixel Styles in the 1st and 3rd rows, while preserving the structural semantic consistency of the content image. In contrast, baseline methods achieve some stylization but suffer from issues like insufficient stylization and style semantic leakage. 
Figure~\ref{fig:Zoomed-compare} presents additional comparisons of scaling details.
We also compared our approach with the mask-based method SCSA~\cite{shang2025scsa}. Given SCSA's mask constraints, we benchmark on pre-computed results from its public repository as shown in Figure~\ref{fig:scsa}.

\textbf{Quantitative Comparison.}
We evaluate our method through comparisons with state-of-the-art approaches, including CNN-based, Transformer-based, and Diffusion-based methods. All baselines are adapted from their original settings to support an arbitrary number of reference images. We rely on public implementations with recommended configurations, and keep the content image and style map inputs for each generated result are consistent. The results of our quantitative analysis are shown in Table~\ref{tab:quantity-comparison}.

\begin{table}[t]
    \centering
    \scalebox{0.8}{
        \setlength{\tabcolsep}{7pt}
        \renewcommand{\arraystretch}{1.2}
        \small
        \begin{tabular}{cccc}
            \toprule
            Configuration & LPIPS~\cite{Zhang2018TheUE} ↓ & FID~\cite{heusel2017gans} ↓ & Style ↑ \\
            \midrule
            Ours ($\lambda_c=0.26$, default) & 0.3716 & \textbf{16.89} & \textbf{0.5337}\\
            Ours ($\lambda_c=0.29$, above) & \textbf{0.3689} & 27.26 & 0.4172 \\
            Ours ($\lambda_c=0.22$, below) & 0.4354 & 21.84 & 0.4562 \\
            \midrule
            w/o RSL & 0.5195 & 23.56 & 0.4387\\
            w/o GCL & 0.6822 & 30.83 &  0.4150\\
            \bottomrule
        \end{tabular}
    } 
    \caption{Ablation study on our proposed components.}
    \label{tab:quan-ablation}
\end{table}

\begin{figure*}[!htbp]
    \centering
    \includegraphics[width=1.0\linewidth]{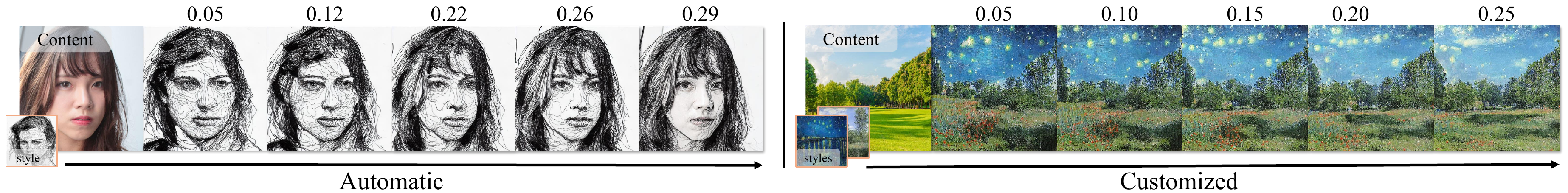}
    \caption{Visual impact of the content loss weight $\lambda_c$. As the global content loss rises from left to right, generated results better preserve the content image's structure while keeping stylization. In automatic mode, we set 0.26 as the default; in custom mode, this weight can be selected by the user.}
    \label{fig:ab-2}
\end{figure*}

\begin{figure}[t]
    \centering
    \includegraphics[width=1.0\linewidth]{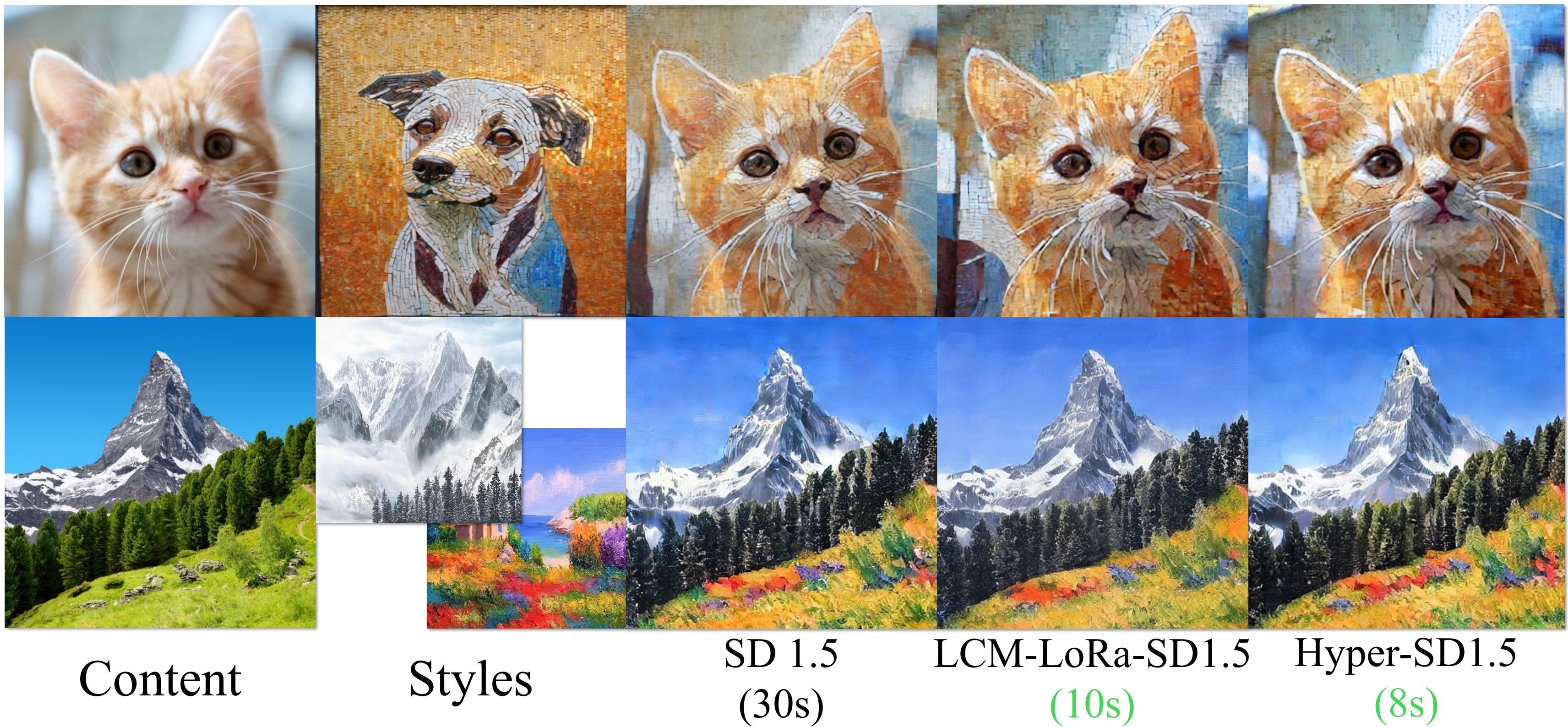}
    \caption{Compatibility analysis. We integrated with accelerated models reduces optimization steps from 150 to 28. The reduced inference times (highlighted in green) demonstrate a significant speedup while maintaining high-fidelity stylization quality.}
    \label{fig:acce}
\end{figure}

\subsection{Ablation Studies} 
To evaluate the effectiveness of our components and the appropriateness of the hyperparameter settings, we conduct ablation studies in both quantitative and qualitative ways.
As shown in Fig.~\ref{fig:ab-1} (left), the global content loss alone preserves semantic structure but yields weak stylization, whereas the region style loss alone boosts style similarity at the cost of content fidelity. Combining both produces the best trade-off—strong regional stylization with minimal semantic leakage.
In the right panel, removing masks causes leakage (e.g., mountain textures bleeding into grass or dark artifacts polluting the background). Masked transfer confines stylization to matched regions and constrains attention so that queries $(Q)$ do not attend to mismatched keys $(K)$, thereby reducing leakage and confirming the effectiveness of the mask.

Figure~\ref{fig:ab-2} shows the impact of content loss weight $\lambda_c$ on the generated results. As $\lambda_c$ increases from left to right, the generated image's semantic structure aligns more closely with the original while maintaining the stylized effect. 
We also conduct ablation studies on our proposed components in quantitative ways as shown in Table~\ref{tab:quan-ablation}.

\subsection{Compatibility with Accelerated Models}
To further explore the extensibility of our framework, we try to integrate advanced acceleration models (eg., Latent Consistency Model (LCM)~\cite{Luo2023LatentCM} and Hyper-SD~\cite{ren2024hyper}) into our original framework.
As shown in Figure~\ref{fig:acce}, this integration significantly reduces the number of optimization steps from 150 to 28 and cuts the inference time from $\approx$30s to $\approx$8s without much performance change.

\subsection{Robustness Analysis}
\begin{figure}[t]
    \centering
    \includegraphics[width=1.0\linewidth]{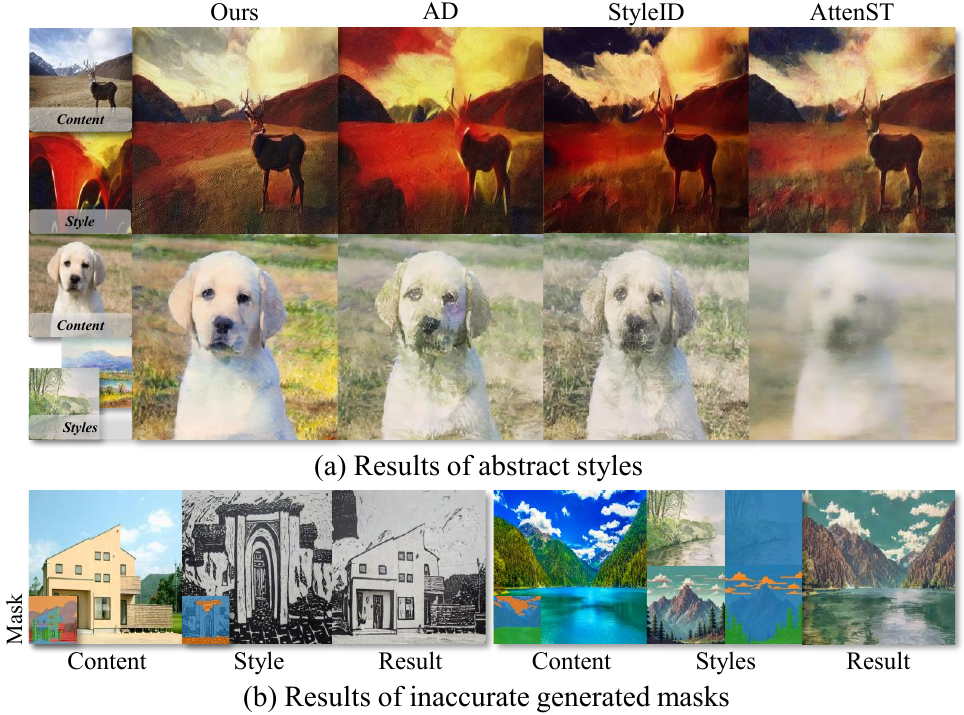}
    \caption{Results of robustness analysis.}
    \label{fig:ab-style}
\end{figure}

To evaluate the robustness of our method, we categorize unexpected inputs into two challenging scenarios: 

(1) Highly abstract style images: As shown in Figure~\ref{fig:ab-style} (a), even with highly abstract style inputs containing minimal semantic cues, our method produces plausible results while preserving the original content, mitigating semantic leakage. These cases highlight potential areas for improvement, such as enhancing the method's sensitivity to subtle or vague style cues, which we consider a direction for future research and refinement.

(2) Inaccurate generated semantic masks: When inputs are semantically ambiguous or structurally complex, this may result in the automatic generated semantic mask lacking sufficient accuracy, thereby degrading the quality of the generated output (shown in Figure~\ref{fig:ab-style} (b)). To mitigate this issue, we support user-provided semantic masks or external models (eg., SAM~\cite{kirillov2023segment}) as initial base masks for subsequent mask optimization, improving mask accuracy and downstream stylization.

We have also introduced an interactive Gradio page that allows users to select their preferred styles for migration and visualize the corresponding stylization results. 
This interface supports exploring different style references and settings in a user-friendly manner, further demonstrating the practicality of our framework in real-world usage. It expands the applicability of our approach and provides a concrete basis for future research on interactive and controllable style transfer; detailed information can be found in the supplementary material.

\section{Conclusions and Limitations}
We present StyleGallery, a training-free framework for arbitrary style transfer that enables region-based, fine-grained control, yielding semantically-aware and interpretable results. It targets current limitations in semantic consistency and user customization, with core technical designs including region clustering via diffusion features, multi-dimensional cluster matching, and a style optimization process that applies fine-grained style loss to semantically relevant regions. Extensive experiments validate StyleGallery's superior quantitative and qualitative performance, paving the way for personalized style customization.

\textit{Limitations.}
Incorrect cluster masks can cause partially broken stylization, which could be addressed by improved clustering or interactive selection. Future work will focus on improving robustness across complex scenarios.

\section*{Acknowledgments}
This work was supported by the Key Program of Hunan Provincial Natural Science Foundation under Grant 2026JJ30028 and the National Natural Science Foundation of China under Grants 62402171, 62472434.
{
    \small
    \bibliographystyle{ieeenat_fullname}
    \bibliography{main}
}
\appendix
\setcounter{figure}{0}
\setcounter{table}{0}
\setcounter{equation}{0}
\clearpage
\setcounter{page}{1}

\section{Statistical Analysis for Style Dataset}
We present an extensive analysis of the contributed style dataset, which encompasses 8 distinct art forms from various periods in Europe, 3 collections of works by renowned painters (Van Gogh, Monet, Zhang Daqian), 4 categories of ancient Chinese art (including Dunhuang murals and ink paintings), and 10 contemporary art styles (such as anime, stick figures, and street art).

\begin{figure}[ht]
    \centering
    \includegraphics[width=0.7\linewidth]{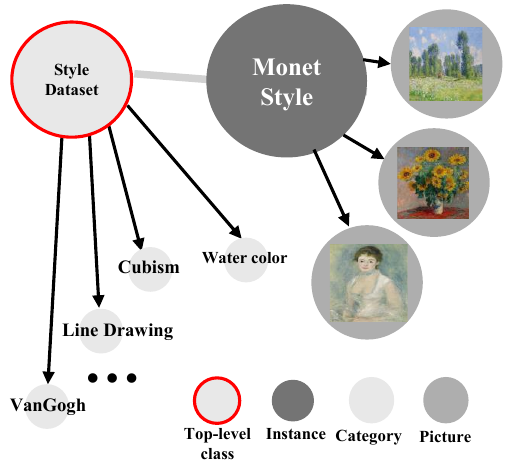}
    \caption{Overall composition of the style dataset.}
    \label{fig:data}
\end{figure}

\begin{figure}[t]
    \centering
    \includegraphics[width=0.9\linewidth]{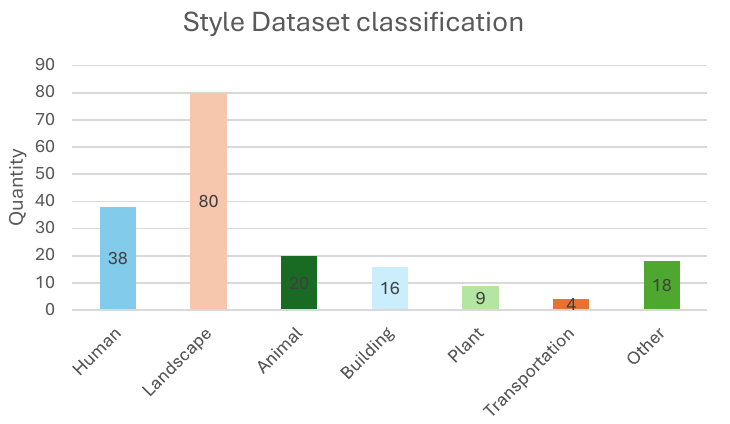}
    \caption{Category distribution in style dataset.}
    \label{fig:dataset-analyse}
\end{figure}

Figure~\ref{fig:data} illustrates the overall composition of our style dataset, while we have conducted a thorough analysis of the distribution of image categories within it. To ensure the diversity and robustness of our style transfer results, our dataset includes images classified into 7 different categories, such as portraits, stylistic paintings, and architectural works. As depicted in Figure~\ref{fig:dataset-analyse}, the top three categories are Landscape (80), Human (38), and Animal (20). In contrast, the Transportation category is represented with a score of only 4, significantly lower than the leading three categories. This disparity arises from the fact that transportation subjects are not frequently represented in traditional paintings.


\section{Cluster Classification} 
\begin{figure}[t]
    \centering
    \includegraphics[width=1.0\linewidth]{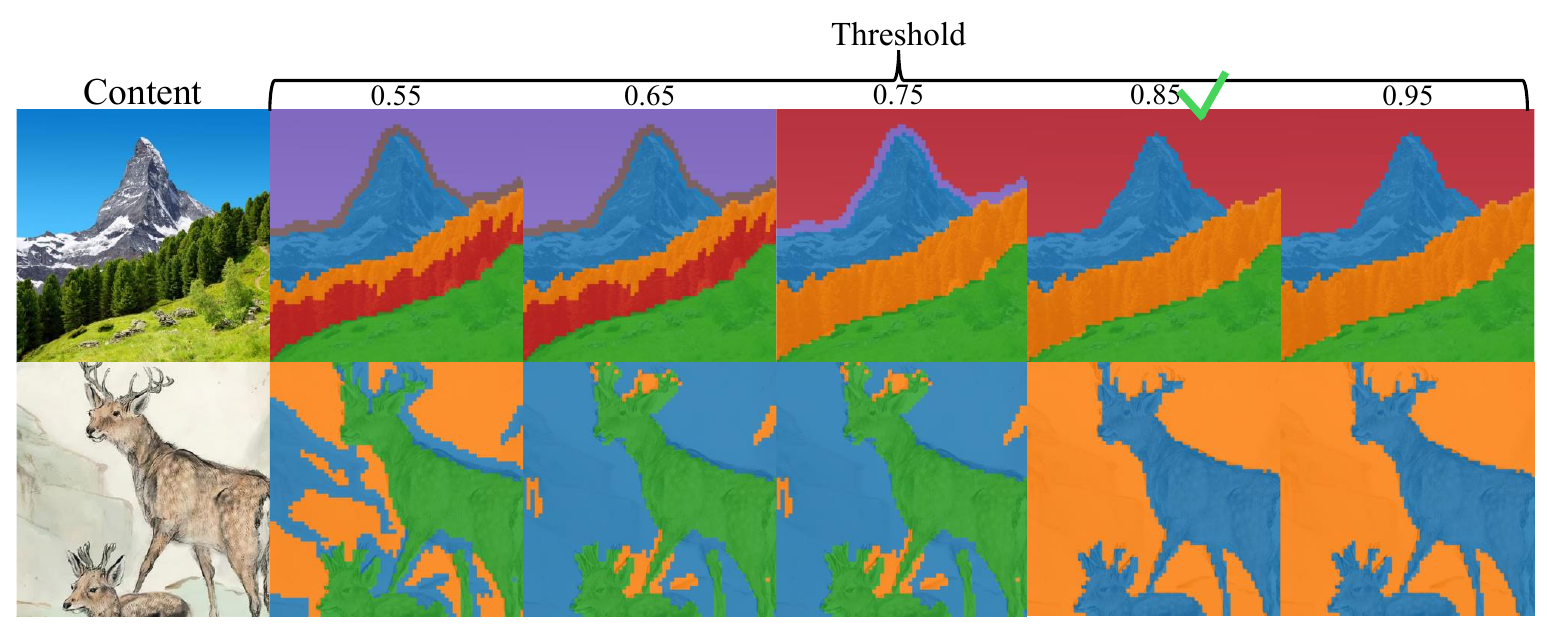}
    \caption{Ablations on cluster merge threshold.}
    \label{fig:sup-thresh}
\end{figure}

We conducted a qualitative analysis of the threshold mentioned in main paper, as shown in Figure~\ref{fig:sup-thresh}, it can be seen that when the semantic fusion threshold is 0.85, the accuracy of semantic region classification is highest.

We also measured the semantic classification accuracy of 20 images across different categories at various thresholds, as shown in Table~\ref{tab:ab-thresh}. For each image, the number of expected semantic regions was manually determined. Correctly identified semantic regions were recorded as +1, while incorrect ones were recorded as 0. Thus, the semantic classification accuracy for each image can be expressed as: 
\begin{equation}
\text{Accuracy} = \frac{\sum_{i=1}^{E} \mathbb{I}(\text{region}_i \text{ is correct})}{E} * \text{100\%}
\end{equation}
where $E$ is the manually determined number of expected semantic regions, and $\mathbb{I}(\cdot)$ is the indicator function that returns $1$ if the region is correctly identified and $0$ otherwise.

Considering the differences between various inputs, we modified the use of depth features and external model as an optional approach. 
Extensive experiments demonstrate that deep features can be effectively utilized as constraints to generate accurate semantic masks.
Furthermore, our method also supports using SAM~\cite{kirillov2023segment} to generate initial base masks, which can subsequently be processed through our mask optimization algorithm to yield accurate semantic masks.

\begin{table}[t]
    \centering
    \small
    \setlength{\tabcolsep}{5pt}
    \renewcommand{\arraystretch}{1.2}
    \scalebox{1.0}{
    \begin{tabular}{lccccccc}
        \toprule
        Threshold & 0.55 & 0.65 & 0.75 & 0.85 & 0.95\\
        \midrule
        \multicolumn{1}{c}{Accuracy $\approx$ ($\%$) ↑} & 52 & 65 & 77 & \textbf{93} & 90\\
        \bottomrule
    \end{tabular}}
    \caption{Ablation study: Effect of merge threshold on semantic classification accuracy. The best performance (93\%) is achieved at threshold 0.85.}
    \label{tab:ab-thresh}
\end{table}

\begin{table*}[ht]
    \centering
    \small
    \setlength{\tabcolsep}{4pt}
    \renewcommand{\arraystretch}{0.9}
    \begin{tabular}{l|ccc|c|c|c}
        \toprule
        Metric & $\lambda_1$=0.05 & $\lambda_1$=0.25 & $\lambda_1$=0.85 & only $\lambda_2$  & $\lambda_2+\lambda_1$ & $\lambda_2+\lambda_1+\lambda_3$\\
        \midrule
        LPIPS~\cite{Zhang2018TheUE} ↓ & 0.2345 & \textbf{0.2317} & 0.2323 & 0.2413 & 0.2321 & 0.2317\\
        Accuracy $\approx$ ($\%$) ↑ & 86 & \textbf{89} & 87 & 87 & 89 & 89\\
        \bottomrule
    \end{tabular}
    \caption{Ablation study on similarity weight. It shows that values of $\lambda_1$ (low or high) may yield erroneous semantic correspondences, and indicate that relying solely on semantic features extracted by DINOv2~\cite{Oquab2023DINOv2LR} can lead to mismatches, whereas incorporating statistical features further improves matching accuracy.}
    \label{tab:ab-simi}
\end{table*}

\section{Cluster Matching} 
\begin{figure}[t]
    \centering
    \includegraphics[width=1.0\linewidth]{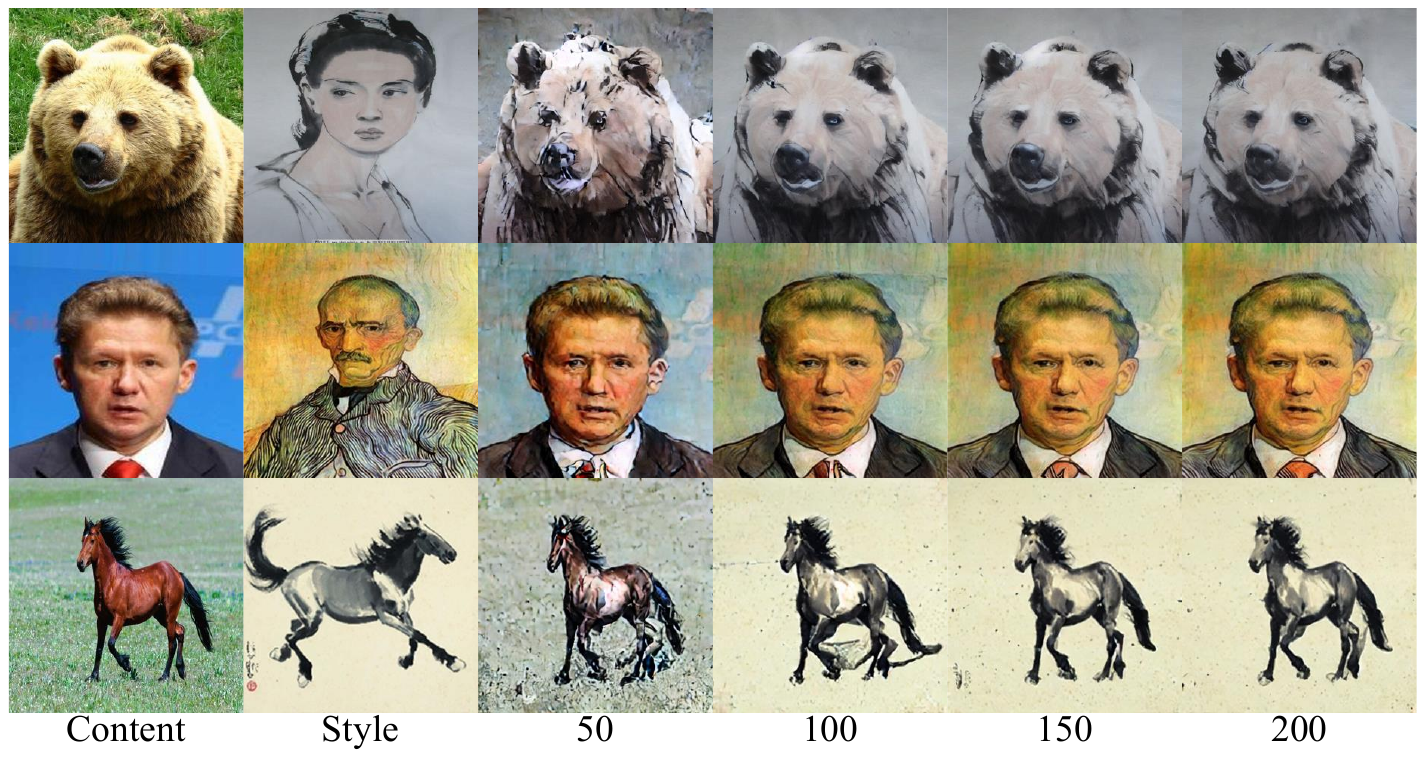}
    \caption{Ablations on optimization steps.}
    \label{fig:sup-op}
\end{figure}

We validate the reasonableness of the weight ratios across three dimensions during semantic matching. 
The weight $\lambda_2$ for semantic similarity was fixed at 1.0. Considering that positional similarity serves only as an auxiliary factor, its weight $\lambda_3$ was set to 0.125.
We randomly generated 50 sets of results and performed quantitative analysis on the weight sets, and manually calculated the accuracy rate of semantic region matching shown in Table~\ref{tab:ab-simi}. 
We also tested the results of style transfer under different optimization time steps, as shown in Figure~\ref{fig:sup-op}.

\section{Details of User Study} 
To validate our qualitative analysis, we conducted a user study on 10 style transfer results, posing three questions. Each question presented three results: one from our method and two from state-of-the-art methods, StyleID~\cite{Chung2023StyleII} and AD~\cite{zhou2025attention}. Users rated each result anonymously on a scale of 1 to 5 based on the following criteria:
1): Visual: Alignment of visual style with the style image set;
2): Region: Accuracy and quality of semantic region style transfer;
3): Content: Similarity of the generated image's semantic information and spatial layout to the source content image.

We gathered results from 50 participants, with 76\% having computer science background (CS) and 24\% not (Non-CS). The final scoring results are as shown in Table~\ref{tab:user-study}. Our method shows better performance on all criteria. 

\begin{table}[t]
    \centering
    \small 
    \setlength{\tabcolsep}{5pt} 
    \renewcommand{\arraystretch}{0.7} 
    \begin{tabular}{lccc} 
        \toprule
        Metric & StyleID & AD & Ours \\
        \midrule
        Visual  & 3.76 & 3.77 & \textbf{4.43} \\
        Region  & 3.44 & 3.75 & \textbf{4.41} \\
        Content & 4.35 & 3.73 & \textbf{4.46} \\
        \bottomrule
    \end{tabular}
    \caption{User preference scores across three aspects.}
    \label{tab:user-study}
\end{table}

\begin{table}[t]
    \centering
    \small
    \setlength{\tabcolsep}{2pt}
    \scalebox{0.7}{%
    \begin{tabularx}{0.9\linewidth}{l *{1}{>{\centering\arraybackslash}X}}
        \toprule
        & Distribution (\%) \\
        \bottomrule
        CS     & 76 \\
        Non-CS & 24 \\
        \bottomrule
    \end{tabularx}%
    }
    \caption{User distribution.}
    \label{tab:User Distribution}
\end{table}

\begin{figure}[t]
    \centering
    \includegraphics[width=0.8\linewidth]{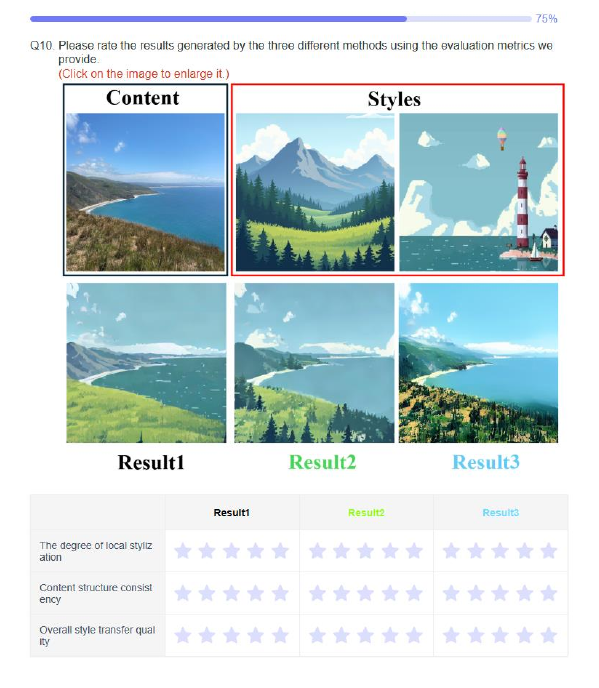}
    \caption{User study interface.}
    \label{fig:interface}
\end{figure}

To further emphasize the rigor of our user preference research, we categorized participants in the user preference survey into two groups: those with computer science backgrounds (CS) and those without (Non-CS), as illustrated in Table~\ref{tab:User Distribution}. We also calculated preference scores for each group, as shown in Table~\ref{tab:user-com}. As noted in Table~\ref{tab:User Distribution}, participants without a computer science background tend to prioritize first impressions, resulting in improved overall visual and local stylization scores for StyleID~\cite{Chung2023StyleII}. Conversely, AD~\cite{zhou2025attention}, due to its content leakage issue, may yield generated images that significantly differ from the original content. Consequently, for participants without a computer science background, AD's scores declined across all three metrics. Despite a decrease in scores, our method continues to achieve the highest ratings among both groups. The user interface is depicted in Figure~\ref{fig:interface}.

\begin{table}[t]
\centering
\small
\setlength{\tabcolsep}{2pt}
\begin{tabularx}{0.48\linewidth}{l *{3}{>{\centering\arraybackslash}X}}
    \toprule
    Metric & StyleID & AD & \textbf{Ours}\\
    \midrule
    Visual & 3.75 & 3.87 & \textbf{4.50} \\
    Region & 3.36 & 3.85 & \textbf{4.60} \\
    Content & 4.40 & 3.81 & \textbf{4.54} \\
    \bottomrule
\end{tabularx}
\hfill
\begin{tabularx}{0.48\linewidth}{l *{3}{>{\centering\arraybackslash}X}}
    \toprule
    Metric & StyleID & AD & \textbf{Ours}\\
    \midrule
    Visual & 3.78 & 3.44 & \textbf{4.20} \\
    Region & 3.69 & 3.45 & \textbf{3.80} \\
    Content & 4.20 & 3.46 & \textbf{4.22} \\
    \bottomrule
\end{tabularx}
\caption{Left: CS background users; Right: non-CS background users.}
\label{tab:user-com}
\end{table}

\section{Additional Experiments}
Here we provide more experimental results including user personalization gradio interface, automatic generated style transfer results, user-specific style transfer results, comparisons with existing state-of-the-art large models.

\subsection{Interface Gradio}
\begin{figure*}[t]
    \centering
    \includegraphics[width=0.79\linewidth]{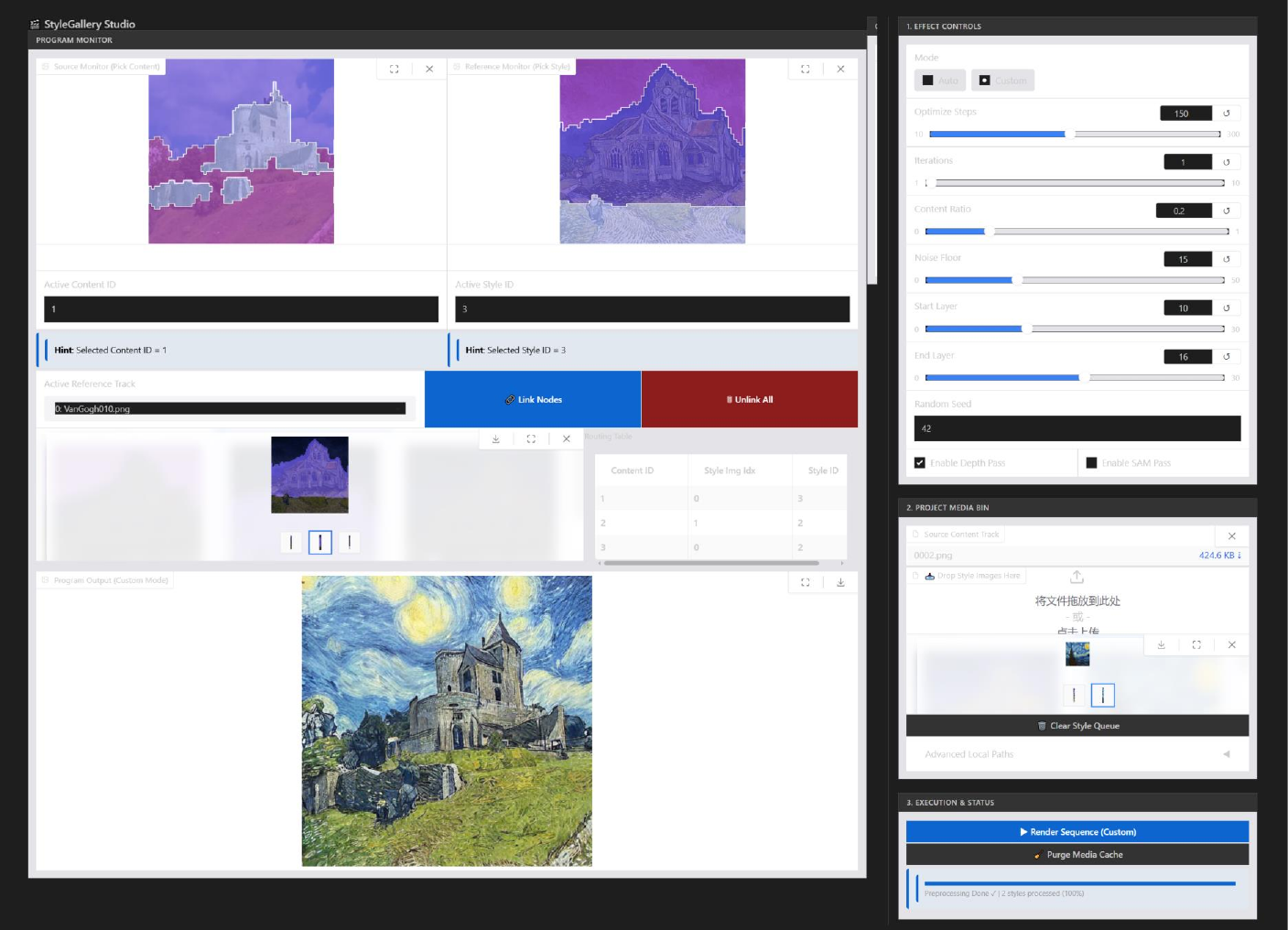}
    \caption{User personalization gradio interface.}
    \label{fig:sup-gradio}
\end{figure*}

As shown in Figure~\ref{fig:sup-gradio}, We create a Gradio page supporting two modes: Auto and Custom. In Custom mode, users can select desired parameters by clicking or sliding, and create personalized styles by adding matching relationships between clusters.

\subsection{Different Backbones}
\begin{figure*}[ht]
    \centering
    \includegraphics[width=0.8\linewidth]{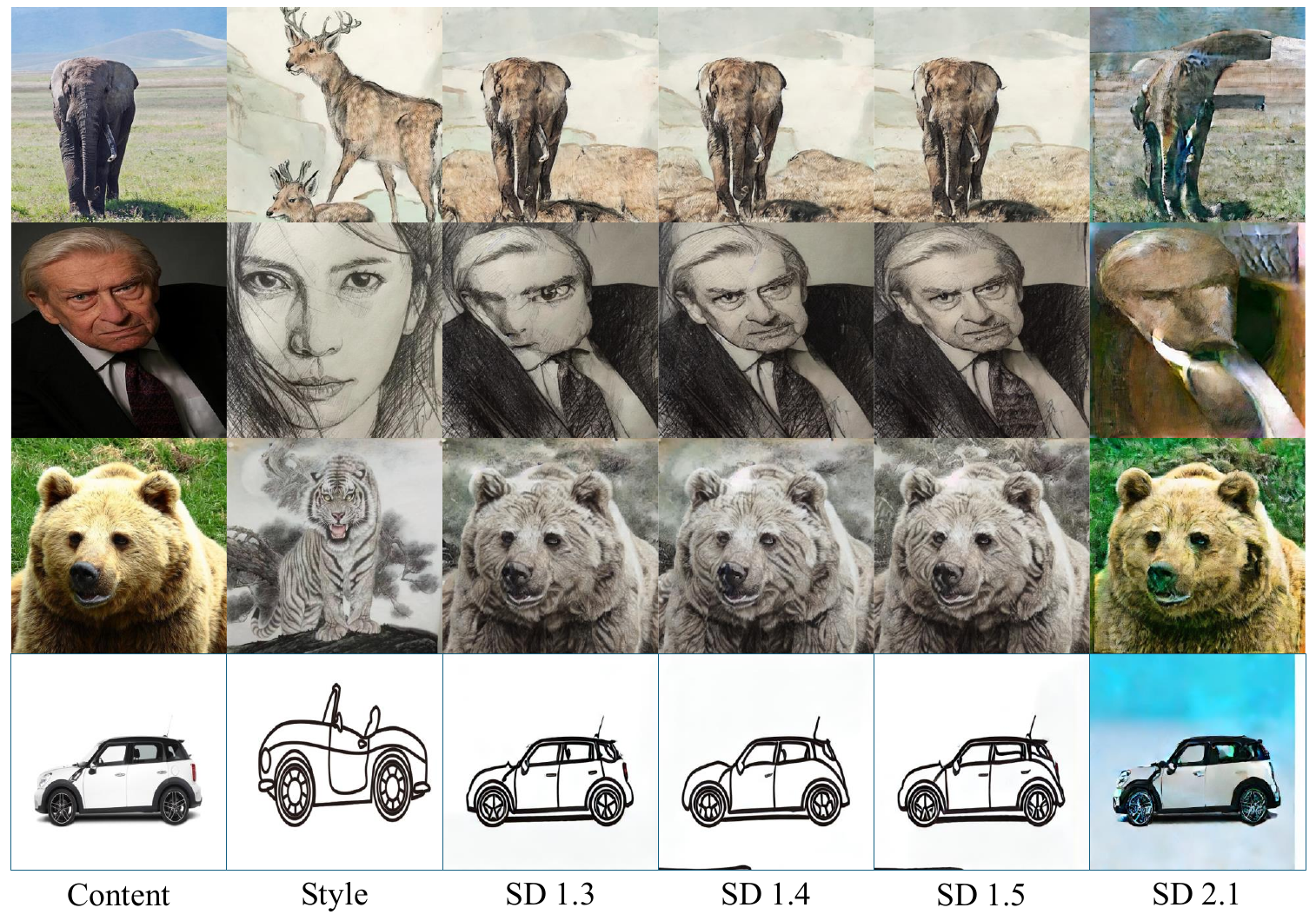}
    \caption{Generation results under different backbones.}
    \label{fig:sup-backbone}
\end{figure*}

Currently, most style transfer methods based on diffusion models utilize Stable Diffusion 1.5~\cite{rombach2022high}. This naturally raises the question: would using different versions of the Stable Diffusion model yield different results? Therefore, we conducted tests on SD1.3, SD 1.4, SD 1.5 and SD 2.1. The results are shown in Figure~\ref{fig:sup-backbone}.
\subsection{More Results}
\begin{figure*}[t]
    \centering
    \includegraphics[width=0.8\linewidth]{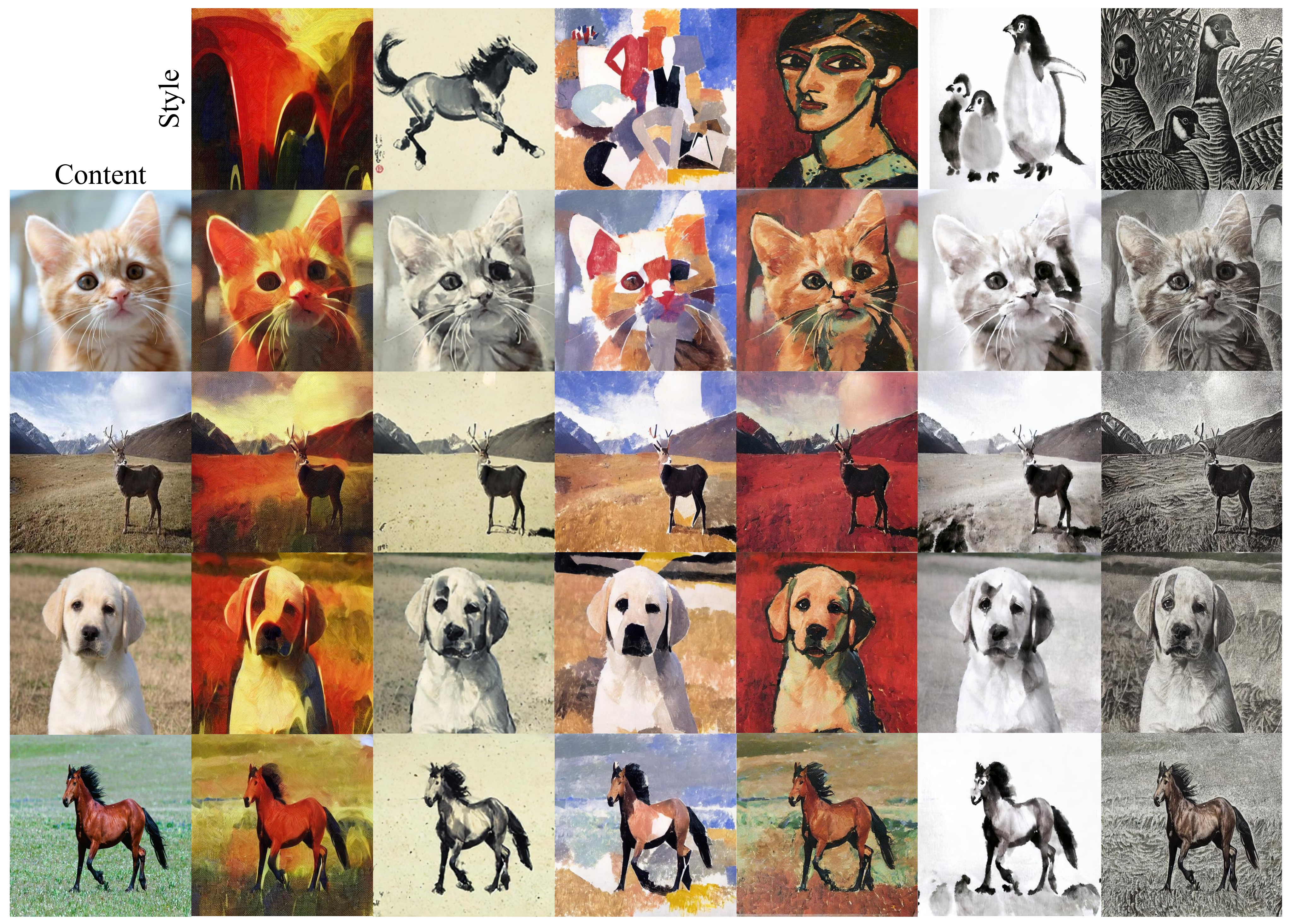}
    \caption{More style transfer results.}
    \label{fig:r1}
\end{figure*}

\begin{figure*}[t]
    \centering
    \includegraphics[width=0.8\linewidth]{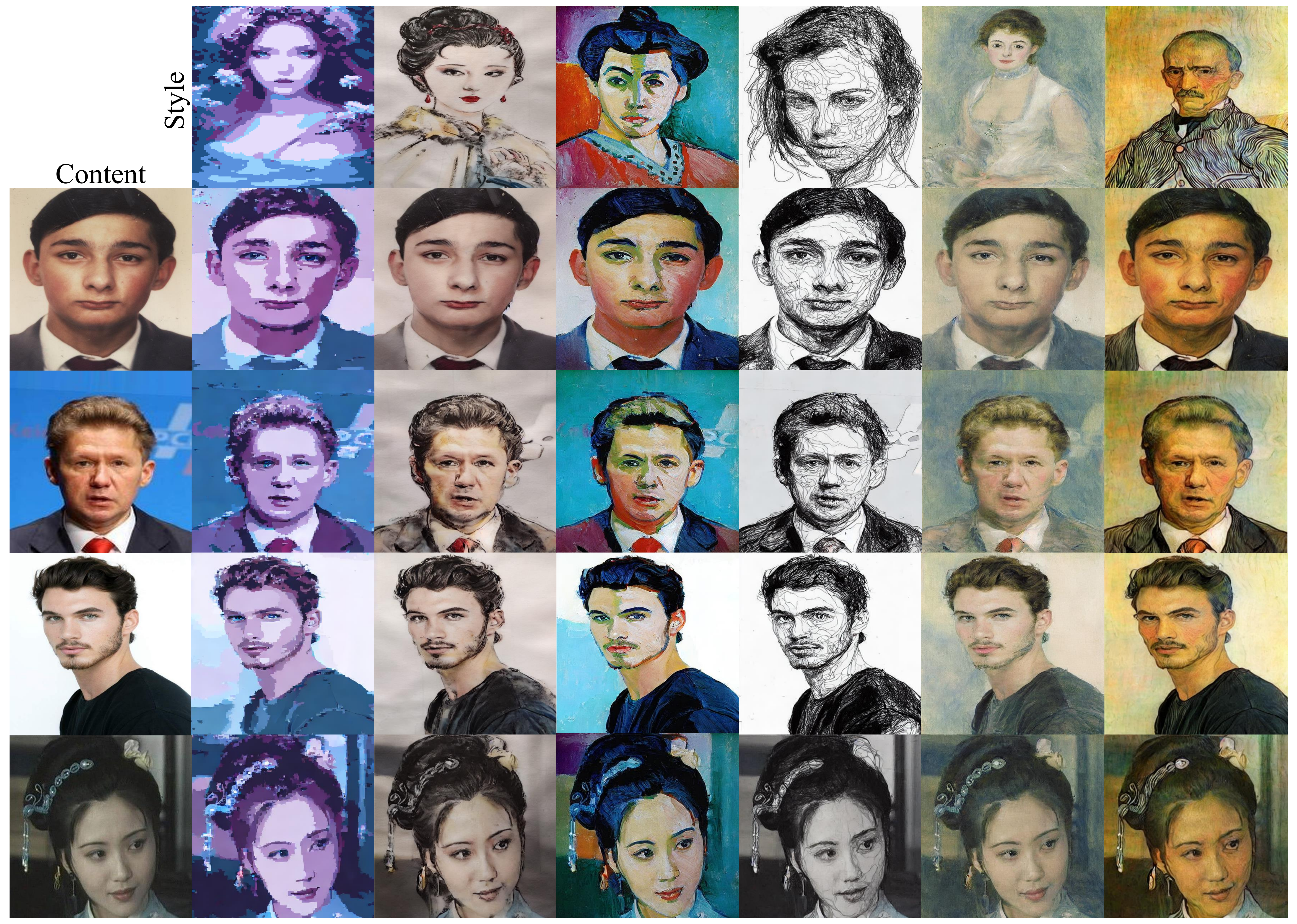}
    \caption{More style transfer results.}
    \label{fig:r2}
\end{figure*}

\begin{figure*}[t]
    \centering
    \includegraphics[width=0.8\linewidth]{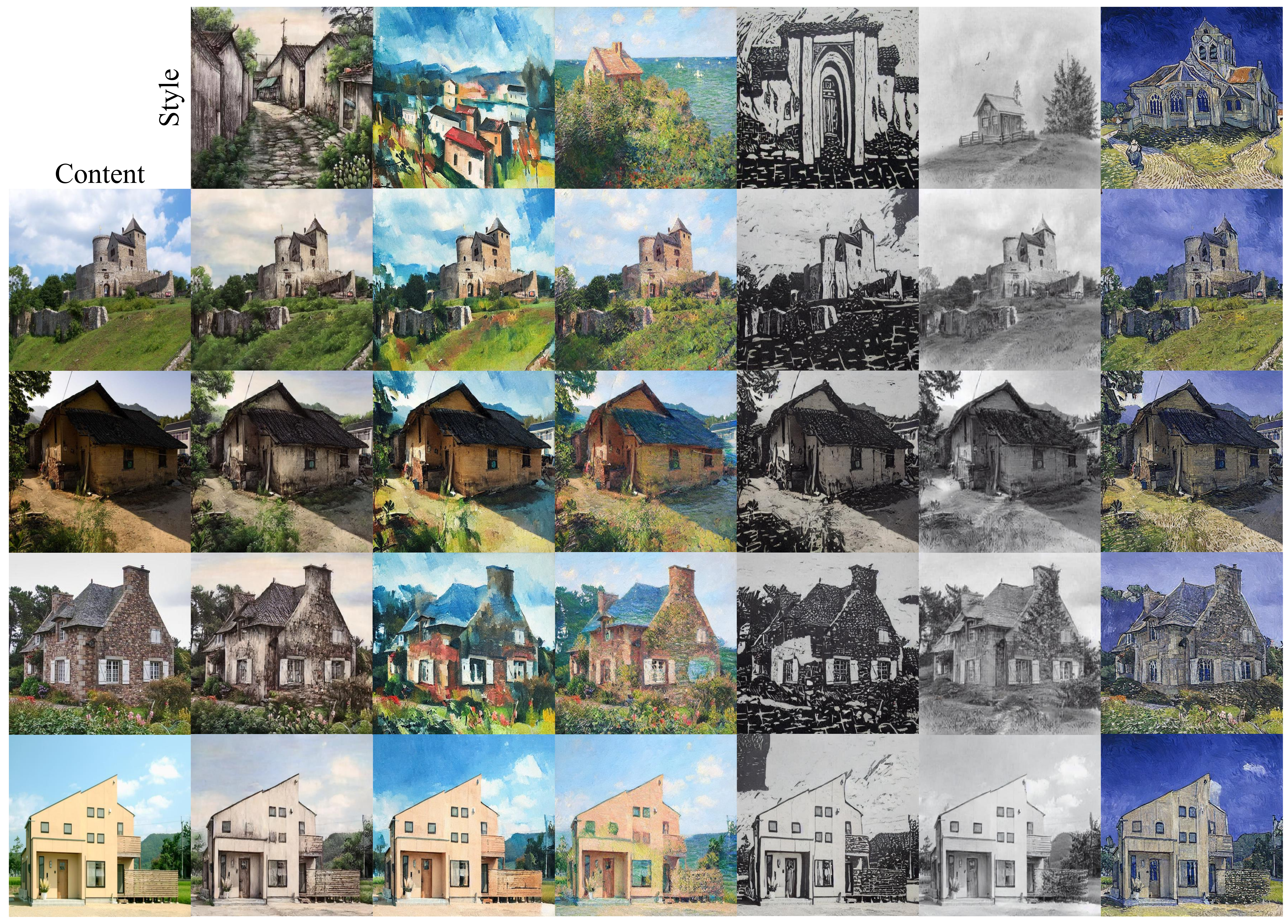}
    \caption{More style transfer results.}
    \label{fig:r3}
\end{figure*}

Finally, we provide additional results:

(1) In Figure~\ref{fig:r1}, Figure~\ref{fig:r2} and Figure~\ref{fig:r3}, we display additional results of automatic generated 
results from single reference.

\begin{figure*}[t]
    \centering
    \includegraphics[width=0.9\linewidth]{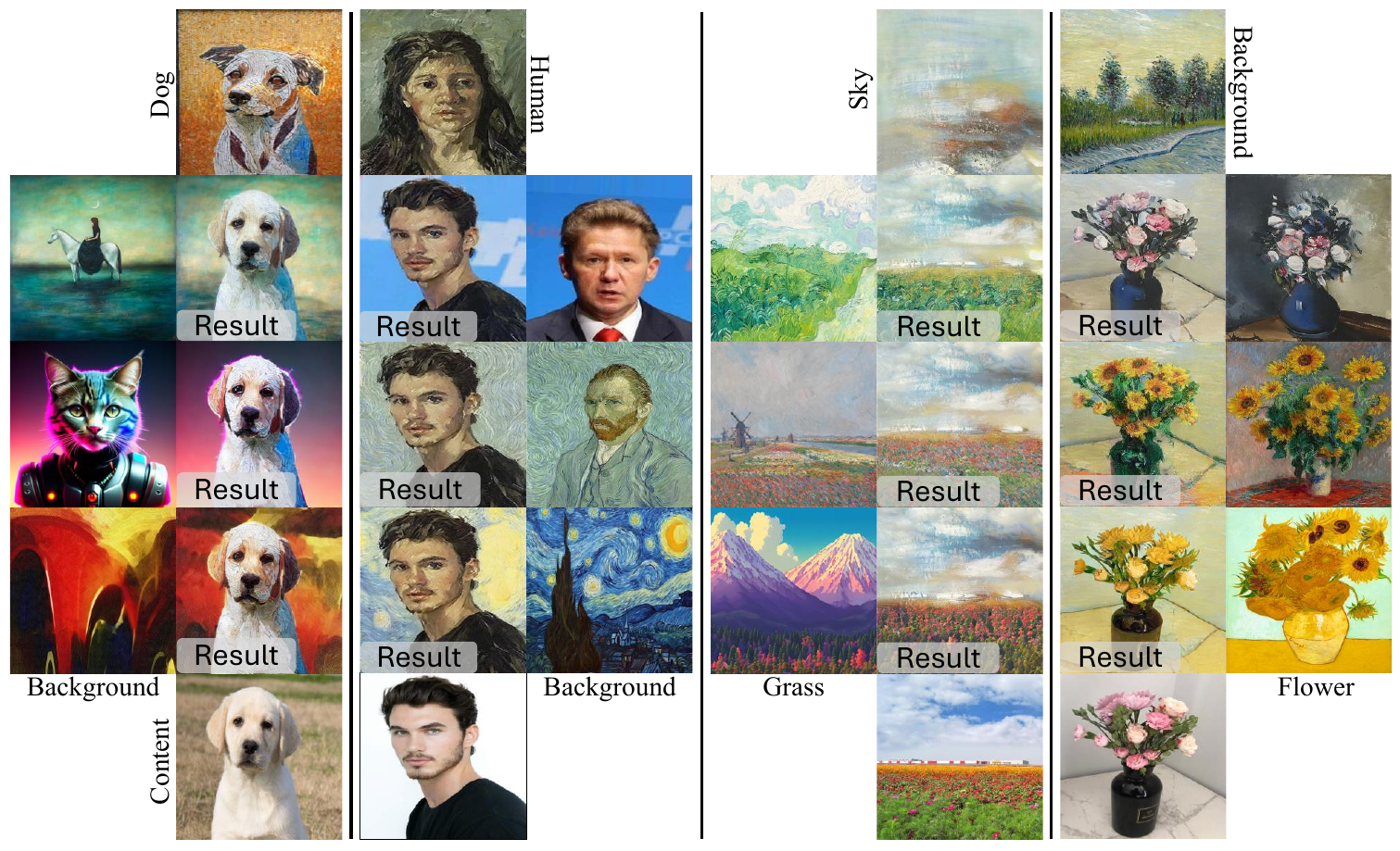}
    \caption{More personalized style transfer results.}
    \label{fig:sup-custom}
\end{figure*}

(2) In Figure~\ref{fig:sup-custom}, we present more personalized style transfer results.

\begin{figure*}[t]
    \centering
    \includegraphics[width=0.9\linewidth]{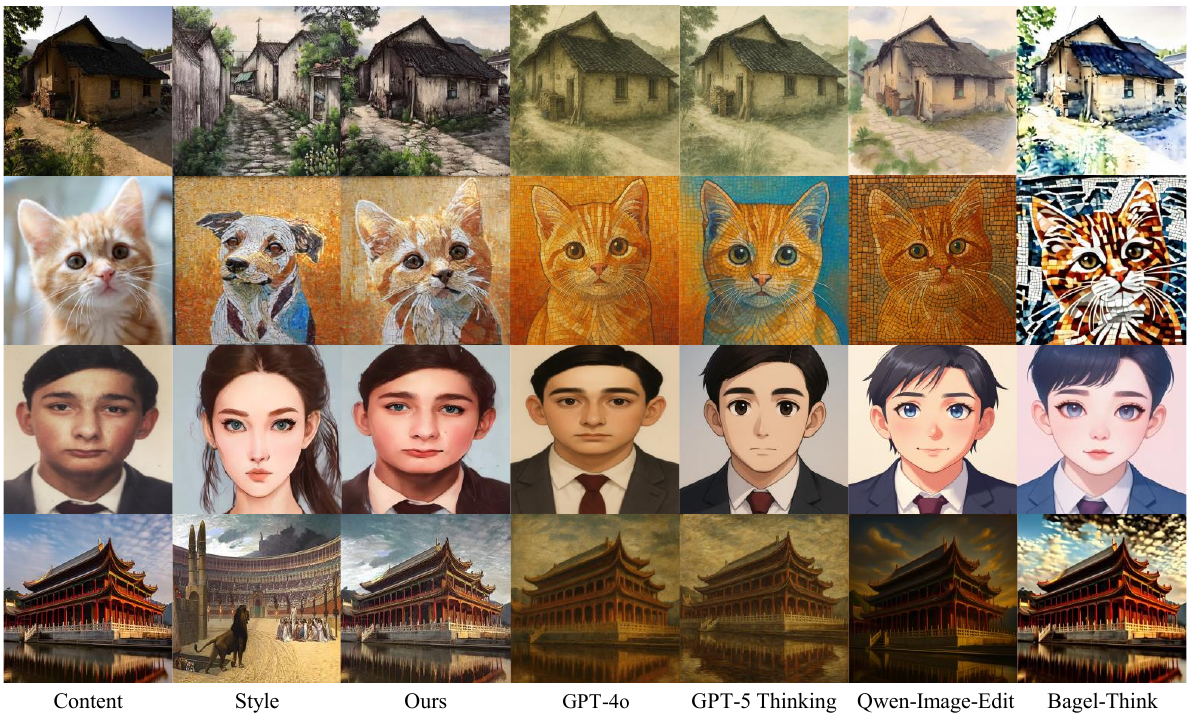}
    \caption{More qualitative comparison results with large models.}
    \label{fig:sup-llm}
\end{figure*}

(3) In Figure~\ref{fig:sup-llm}, we show comparisions with state-of-art large models with image editing capabilities.

\section{Limitations}
To encourage further research in this field, we briefly outline limitation of StyleGallery, which also represent potential directions for future work.

When the semantic structure of an image is overly complex or abstract—such as when a mountain in the foreground obstructs the view of a lake or clouds behind it—we may inadvertently cluster these areas together. Although the attention feature $Q$ can assist in guiding content, the potential for unexpected semantic leakage remains. This represents an inherent challenge in image processing, which could be addressed through more advanced semantic segmentation models or methodologies.

\clearpage


\end{document}